\pdfoutput=1
\documentclass[11pt]{article}

\usepackage{acl}

\usepackage{times}
\usepackage{latexsym}

\usepackage[T1]{fontenc}
\usepackage[utf8]{inputenc}

\usepackage{microtype}

\usepackage{diagbox}
\usepackage{pifont}
\usepackage{tabularx}
\usepackage{todonotes}
\usepackage{booktabs}
\usepackage{diagbox}
\newcommand*\rot{\rotatebox{90}}
\usepackage{adjustbox}
\newcommand{\ts}{\textsuperscript}
\title{NAP at SemEval-2023 Task 3: Is Less Really More? (Back-)Translation as Data Augmentation Strategies for Detecting Persuasion Techniques}

\author{Neele Falk$^*$, Annerose Eichel$^*$, Prisca Piccirilli$^*$ \\
         Institute for Natural Language Processing, University of Stuttgart \\ 
         \texttt{\{firstname.lastname\}@ims.uni-stuttgart.de}}
  \newcommand\nnfootnote[1]{%
  \begin{NoHyper}
  \renewcommand\thefootnote{}\footnote{#1}%
  \addtocounter{footnote}{-1}%
  \end{NoHyper}
}

\begin{document}
\maketitle

\nnfootnote{*All authors contributed equally to this work.}

\begin{abstract}
Persuasion techniques detection in news in a multi-lingual setup is non-trivial and comes with challenges, including little training data. %We present a system which 
Our system successfully leverages (back-)translation as data augmentation strategies with multi-lingual transformer models for the task of detecting persuasion techniques. The automatic and human evaluation of our augmented data allows us to explore whether (back-)translation aid or hinder performance. %w.r.t. the task at hand.
Our in-depth analyses indicate that both data augmentation strategies boost performance; however, balancing human-produced and machine-generated data seems to be crucial.
%\textbf{RQ:} can machine translation be used for data augmentation techniques for the task of detecting persuasion techniques, or does it hinder performance because of wrong label and information transfer. Translation or backtranslation technique better? 
%\textbf{Main finding(s): } 
\end{abstract}

\section{Introduction}
\label{sec:introduction}

The SemEval 2023 Task 3 \cite{semeval2023task3} aims at analyzing online news by detecting \textit{genre}, \textit{framing}, and \textit{persuasion techniques}, i.e., \textit{what} is presented \textit{how} using \textit{which rhetoric means}. 
%While genre detection (Subtask 1) determines whether a news article seeks to report objectively or represents a piece of opinion or satire, framing detection (Subtask 2) classifies the frames used in an article, i.e. the perspective under which an article is presented \cite{card-etal-2015-media,card-etal-2016-analyzing}. 
Persuasion techniques detection (Subtask 3) aims to identify which rhetoric means are used to influence and persuade the reader. 
News articles are provided in 
%the six languages
English, German, French, Italian, Polish, and Russian. To foster the development of language-agnostic systems like our approach, the organizers additionally 
introduce three surprise languages -- Spanish, Greek, and Georgian\footnote{Henceforth, we use the following official identifiers: \textit{en}:English, \textit{fr}:French, \textit{it}:Italian, \textit{ru}:Russian, \textit{ge}:German, \textit{po}:Polish, \textit{es}:Spanish, \textit{el}:Greek, and \textit{ka}:Georgian.} -- with test data only.
%Focusing on the detection of persuasion techniques (Subtask 3), 

We build a system that successfully leverages (back-)translation as data augmentation approaches with multi-lingual transformer models to detect persuasion techniques in all 9 languages. We win the task for \textit{fr}, achieve 2\ts{nd} place for \textit{ge}, \textit{it} and \textit{po}, and 3\ts{rd} place for \textit{es}, \textit{el} and \textit{ka}, while ranking mid-field for \textit{ru} and \textit{en}. 
Our main contribution consists in exploring the extent to which data augmentation via (back-)translation boosts performance for the task at hand. Our findings suggest that \textit{more} data does boost performance, especially for under-represented labels. Our in-depth analyses however show that \textit{less tends to be really more} w.r.t balancing natural vs. augmented data, as \textit{more} (augmented) data can severely hurt performance.
%when it comes to balancing the data to not strongly hurt performance in cases it does not help. 
%% are cases where it does not help always large labels or might there be other reasons?

\begin{figure}[t]
\centering
\includegraphics[width=7.5cm]{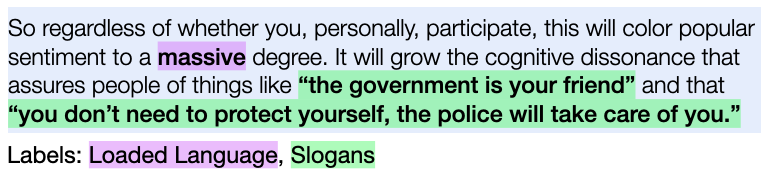}
%\vspace{-0.7cm}
\caption{Example of a multi-labelled paragraph and the corresponding relevant textual spans for the persuasion techniques \textit{Loaded Language} and \textit{Slogans}.} 
\label{fig:example}
\vspace{-0.5cm}
\end{figure}
\section{Background}
\label{sec:background}

%In your own words, summarize important details about the task setup: kind of input and output (give an example if possible); what datasets were used, including language, genre, and size. If there were multiple tracks, say which you participated in.

%Previous shared tasks, including SemEval tasks, have addressed the detection of disinformation and propaganda techniques such as persuasion using automatic systems. \citet{da-san-martino-etal-2019-findings} proposed the \textit{Findings of the NLP4IF-2019 Shared Task on Fine-Grained Propaganda Detection}, asking to detect 18 propaganda techniques in news articles by (i) identifying relevant spans, and (ii) predicting whether a sentence contains propagandist content or not. \citet{da-san-martino-etal-2020-semeval} introduced \textit{SemEval-2020 task 11 on Detection of Persuasion Techniques in News Articles}, where the goal is (i) to determine relevant spans and (ii) predict which of the 14 propaganda technique are used in a given span. \citet{dimitrov-etal-2021-semeval} proposed \textit{SemEval-2021 Task 6: Detection of Persuasion Techniques in Texts and Images}, focusing on identifying an extended set of 22 persuasion techniques given (i) textual or (ii) multi-modal data as well as determining (iii) relevant text spans on document level. 

%Previous shared tasks have addressed the detection of disinformation and propaganda techniques such as persuasion using automatic systems \cite{da-san-martino-etal-2019-findings,da-san-martino-etal-2020-semeval,dimitrov-etal-2021-semeval}. While exact definitions differ, the task of 
Predicting a set of persuasion techniques given a piece of news text in a monolingual setting has been explored in previous %editions of this 
shared tasks \cite{da-san-martino-etal-2019-findings,da-san-martino-etal-2020-semeval,dimitrov-etal-2021-semeval}. Existing successful systems usually include monolingual transformer-based models
%transformer-based models such as BERT, RobERTa, DeBERTa, and XLNet
and ensemble mechanisms to optimally aggregate predictions \cite{mapes-etal-2019-divisive,jurkiewicz-etal-2020-applicaai,chernyavskiy-etal-2020-aschern,tian-etal-2021-mind}. Approaches that additionally focus on provided or external data also show improvement, using techniques such as fine-tuning on additional in-domain data or augmenting the training data \cite{Abujaber-et-al:2021}.

\section{Data Description}
\label{sec:data}

\subsection{Gold Data}
The data consists of news and web articles, covering recent hot topics (such as {\small COVID-19} and climate change) that are multi-lingual (\textit{en, fr, it, ge, ru , po}) and multi-labelled amongst 23 fine-grained persuasion techniques (mapped to 6 coarse-grained categories). %The organizers also provide the
The relevant span-level annotations for each labeled paragraph are also provided. 
Fig. \ref{fig:example} illustrates a multi-labelled instance in \textit{en}, and Table \ref{tab:size-train-data} offers an overview of the training data size for this \textit{gold} dataset. 
This is a rather small and imbalanced dataset regarding both the language in consideration and the labels (cf. Tab.  \ref{tab:label_distribution} in App. \ref{app:data-size}).  
To increase our training data as well to provide additional examples of persuasion techniques for the low represented labels, we use data augmentation techniques.

\subsection{Data Augmentation}
%To avoid using external corpora, 
We experiment with two data augmentation techniques by directly making use of the multi-lingual input that is provided.
First, we generate \textbf{automatic translations} \textit{from} and \textit{to} all possible six languages. Not only does this technique allow us to increase text content, but the transfer of the persuasion techniques along with the corresponding text also increases label representation.  
We also experiment with \textbf{paraphrasing through back-translation}, to augment the data for each language \textit{from} and \textit{to} all possible six languages. As three surprise languages are added in the test set, we also generate (back-)translations to and from these languages, when possible.
Table \ref{tab:langpair-trans-backtrans} provides an overview of the possible combinations we were able to explore depending on the {\small MT} models' availability.
For both techniques, we use the translation system {\small MarianMT}, based on the {\small MarianNMT} framework \cite{Junczys-et-al:2018} and trained using parallel data collected at {\small OPUS} \cite{Tiedemann-Thottingal:2020}. 
The purpose of data augmentation in this work is two-fold. While it enables us to substantially increase our training data size (Tab. \ref{tab:size-train-data}), we are also aware that information contained in the original input can be "lost in (back-)translation" \cite{Troiano-et-al:2020}, which leads to our research question: for the task of detecting persuasion techniques, to which extent can (back-)translation techniques help models' performance? %Besides reporting the models' performance (Section \ref{sec:results}), 
We first conduct automatic and human evaluation on our obtained augmented data.

\begin{table}
\footnotesize
\centering
\begin{tabular}{@{} cl*{6}c @{}}
\toprule
\multicolumn{6}{r}{Source Languages (SL)}
\\\cmidrule{3-8}
& &  en & fr & it & ru & ge & po  \\
\midrule
& en & - & \textcolor{blue}{\ding{51}} \textcolor{red}{\ding{51}} & \textcolor{blue}{\ding{51}} \textcolor{red}{\ding{51}} & \textcolor{blue}{\ding{51}} \textcolor{red}{\ding{51}} & \textcolor{blue}{\ding{51}} \textcolor{red}{\ding{51}} & \textcolor{blue}{\ding{51}} \textcolor{red}{\ding{51}} \\
& fr & \textcolor{blue}{\ding{51}} \textcolor{red}{\ding{51}} & - & \textcolor{blue}{\ding{51}} \textcolor{red}{\ding{55}} & \textcolor{blue}{\ding{51}} \textcolor{red}{\ding{51}} & \textcolor{blue}{\ding{51}} \textcolor{red}{\ding{51}} & \textcolor{blue}{\ding{51}} \textcolor{red}{\ding{51}} \\
& it & \textcolor{blue}{\ding{51}} \textcolor{red}{\ding{51}} & \textcolor{blue}{\ding{55}} \textcolor{red}{\ding{55}} & - & \textcolor{blue}{\ding{55}} \textcolor{red}{\ding{55}} & \textcolor{blue}{\ding{51}} \textcolor{red}{\ding{51}} & \textcolor{blue}{\ding{55}} \textcolor{red}{\ding{55}}   \\
& ru & \textcolor{blue}{\ding{51}} \textcolor{red}{\ding{51}} & \textcolor{blue}{\ding{51}} \textcolor{red}{\ding{51}} & \textcolor{blue}{\ding{55}} \textcolor{red}{\ding{55}} & - & \textcolor{blue}{\ding{55}} \textcolor{red}{\ding{55}} & \textcolor{blue}{\ding{55}} \textcolor{red}{\ding{55}}  \\
& ge & \textcolor{blue}{\ding{51}} \textcolor{red}{\ding{51}} & \textcolor{blue}{\ding{51}} \textcolor{red}{\ding{51}} & \textcolor{blue}{\ding{51}} \textcolor{red}{\ding{51}} & \textcolor{blue}{\ding{55}}\textcolor{red}{\ding{55}} & - & \textcolor{blue}{\ding{51}} \textcolor{red}{\ding{51}}  \\
& po & \textcolor{blue}{\ding{55}} \textcolor{red}{\ding{55}}& \textcolor{blue}{\ding{51}} \textcolor{red}{\ding{51}} & \textcolor{blue}{\ding{55}} \textcolor{red}{\ding{55}} & \textcolor{blue}{\ding{55}} \textcolor{red}{\ding{55}} & \textcolor{blue}{\ding{51}} \textcolor{red}{\ding{51}} & - \\
& es & \textcolor{blue}{\ding{51}} \textcolor{red}{\ding{51}}& \textcolor{blue}{\ding{51}} \textcolor{red}{\ding{51}} & \textcolor{blue}{\ding{51}} \textcolor{red}{\ding{51}} & \textcolor{blue}{\ding{51}} \textcolor{red}{\ding{51}} & \textcolor{blue}{\ding{51}} \textcolor{red}{\ding{51}} &  \textcolor{blue}{\ding{55}} \textcolor{red}{\ding{55}}\\
& el & \textcolor{blue}{\ding{51}} \textcolor{red}{\ding{55}}& \textcolor{blue}{\ding{51}} \textcolor{red}{\ding{51}} & \textcolor{blue}{\ding{55}} \textcolor{red}{\ding{55}} & \textcolor{blue}{\ding{55}} \textcolor{red}{\ding{55}} & \textcolor{blue}{\ding{51}} \textcolor{red}{\ding{55}} & \textcolor{blue}{\ding{55}} \textcolor{red}{\ding{55}} \\
 \rot{\rlap{~Target Languages (TL)}}
& ka & \textcolor{blue}{\ding{55}} \textcolor{red}{\ding{55}}& \textcolor{blue}{\ding{55}} \textcolor{red}{\ding{55}} & \textcolor{blue}{\ding{55}} \textcolor{red}{\ding{55}} & \textcolor{blue}{\ding{55}} \textcolor{red}{\ding{55}} & \textcolor{blue}{\ding{55}} \textcolor{red}{\ding{55}} & \textcolor{blue}{\ding{55}} \textcolor{red}{\ding{55}} \\
\bottomrule 
\end{tabular}
\captionsetup{font=footnotesize}
\caption{Language pairs covered (\ding{51}) and not covered (\ding{55}) by {\small marianMT} models for translation (in \textcolor{blue}{blue}) and back-translation (in \textcolor{red}{red}). The direction of \textit{translation} is from {\small SL} to {\small TL} and back to {\small SL} for \textit{back-translation.}}
\label{tab:langpair-trans-backtrans}
\vspace{+0.17cm}
\end{table}
\begin{table}[ht]
\footnotesize
\centering
\begin{tabular}{cccccc}\toprule
    en & fr & it & ru & ge & po  \\\midrule
    48.06 & 32.80 & 45.79 & 20.84 & 33.34  & 21.23 \\\bottomrule
\end{tabular}
\captionsetup{font=footnotesize}
\caption{4-gram {\small BLEU} score average per language.}
\label{tab:avg_bleu}
%\vspace{-0.3cm}
\end{table}

\begin{table}[b]
%\vspace{-0.3cm}
\footnotesize
\centering
\begin{tabular}{@{} l*{6}r @{}}
\toprule
\multicolumn{5}{r}{Training Datasets}
\\\cmidrule{2-6}
 & gold & +{\small BT}-sl & +{\small T}+{\small BT} & +{\small T}+{\small BT}-sl & +span  \\\midrule
 en & 3,761 & 22,561 & 25,968 & 29,728 & 7,521\\
 fr & 1,694 & 11,852 & 17,700 & 21,086 & 3,387\\
 it & 1,746 & 6,981 & 10,248 & 11,993 & 3,491\\
 ru & 1,246 & 4,981 & 9,189 & 10,434 & 2,491 \\
 ge & 1,253 & 7,513 & 14,691 & 15,943 & 2,505\\
 po & 1,233 & 3,697 & 6,642 & 6,642 & 2,465\\
\textbf{total} & \textbf{10,933} &  \textbf{57,585} & \textbf{84,438} & \textbf{95,826} &  \textbf{21,860}  \\\bottomrule
\end{tabular}
\captionsetup{font=footnotesize}
\caption{Training data size per language. \textit{gold} is the original task data, to which are added all possible (back-)translations (+{\small T} and +{\small BT}) - with or without the surprise languages (\textit{sl}) as pivot languages - and the relevant textual \textit{span}s.}
\label{tab:size-train-data}
%\vspace{-0.3cm}
\end{table}

\begin{figure*}[!htb]
\minipage{0.49\linewidth}  \includegraphics[width=\textwidth]{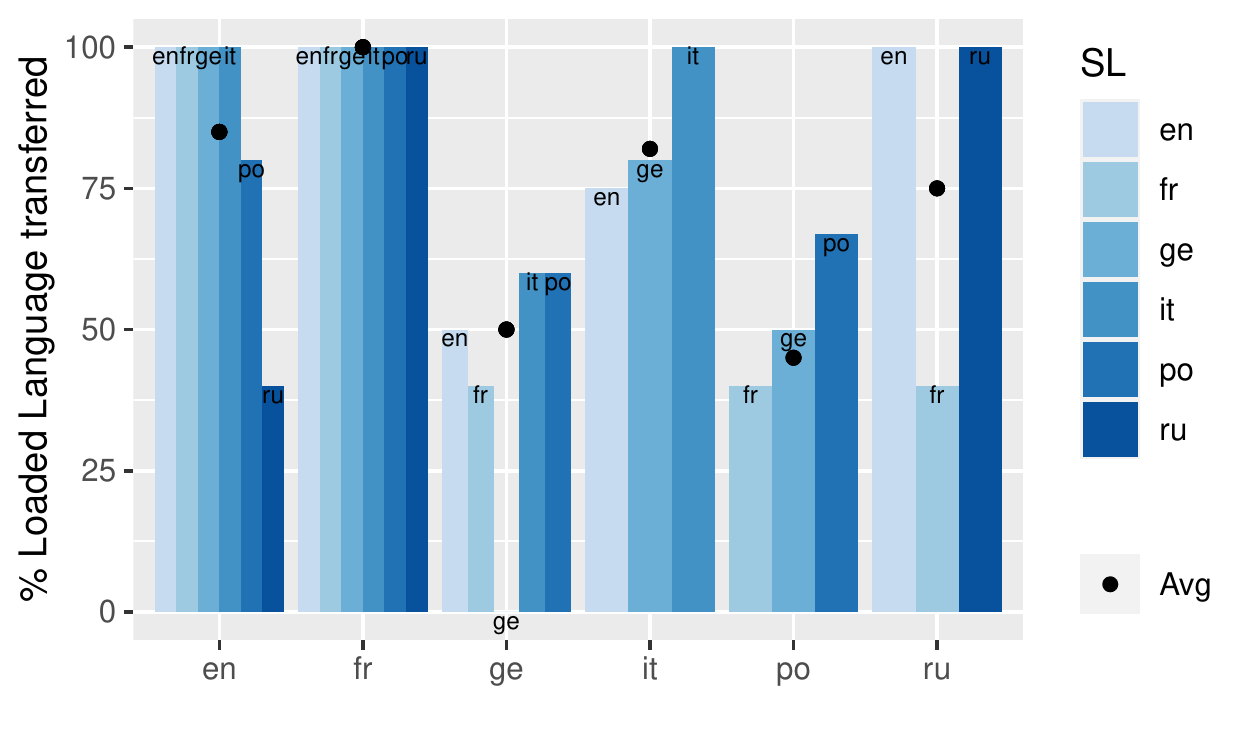}
\vspace{-0.5cm}
\captionsetup{font=footnotesize}
\caption{\% of \textit{translations} where {\small \textit{Loaded Language}} is \textit{transferred} irresp. of ({\small \textbf{Avg}}) and according to the {\small \textbf{SL}}, resp.}
\label{fig:T_loaded_preserved}
\endminipage\hfill
\minipage{0.49\linewidth}  
\includegraphics[width=\textwidth]{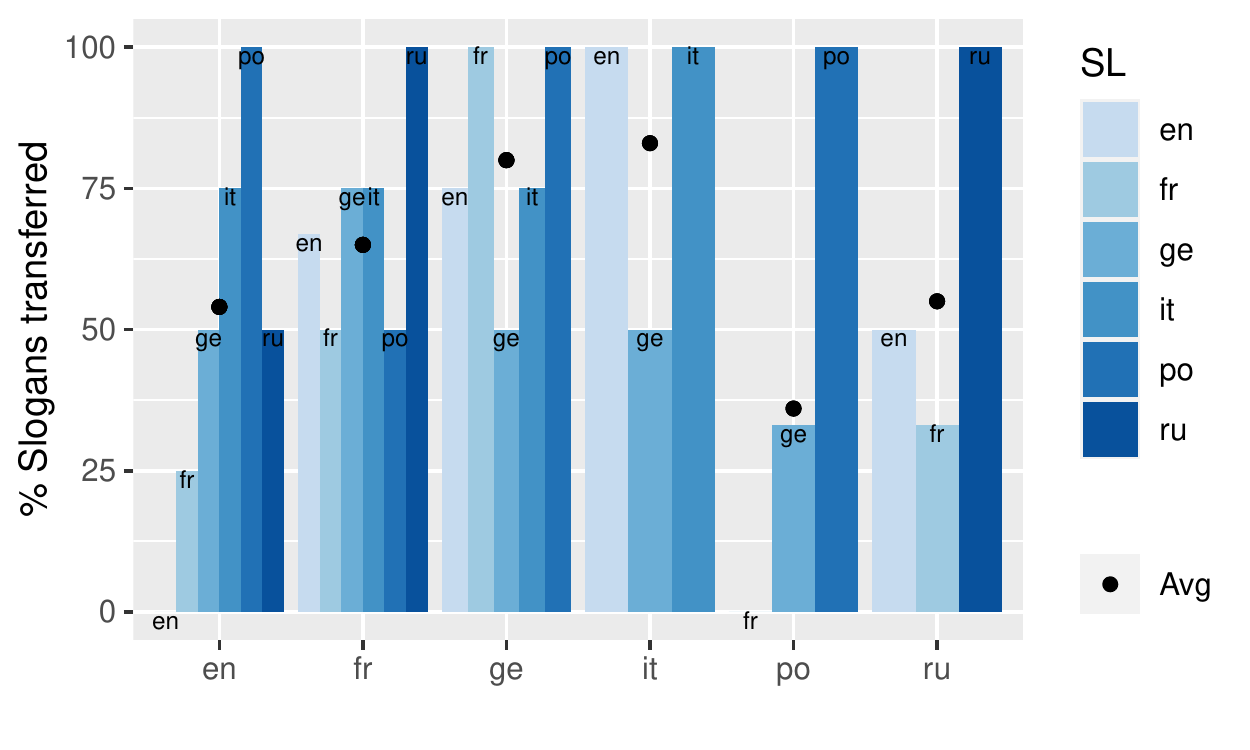}
\vspace{-0.5cm}
\captionsetup{font=footnotesize}
\caption{\% of \textit{translations} where  {\small \textit{Slogans}} is \textit{transferred} irresp. of ({\small \textbf{Avg}}) and according to the {\small \textbf{SL}}, resp.}\label{fig:T_slogans_preserved}
\endminipage
\end{figure*}
\begin{figure*}[htb]
\minipage{0.49\linewidth}  
\includegraphics[width=\textwidth]{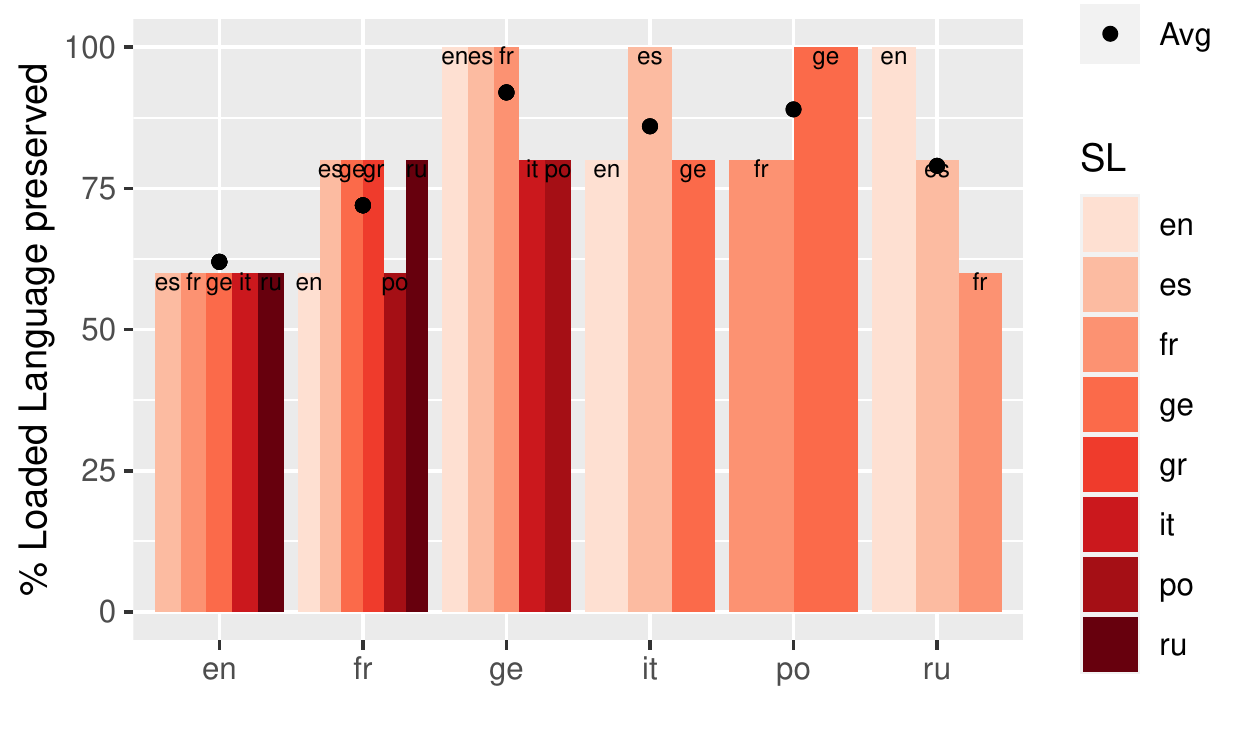}   
\vspace{-0.5cm}
\captionsetup{font=footnotesize}
\caption{\% of \textit{back-translations} where {\small \textit{Loaded Language}} is \textit{preserved} irresp. of ({\small \textbf{Avg}}) and according to the {\small \textbf{SL}}, resp.}\label{fig:BT_loaded_preserved}
\endminipage\hfill
\minipage{0.49\linewidth}
\includegraphics[width=\textwidth]{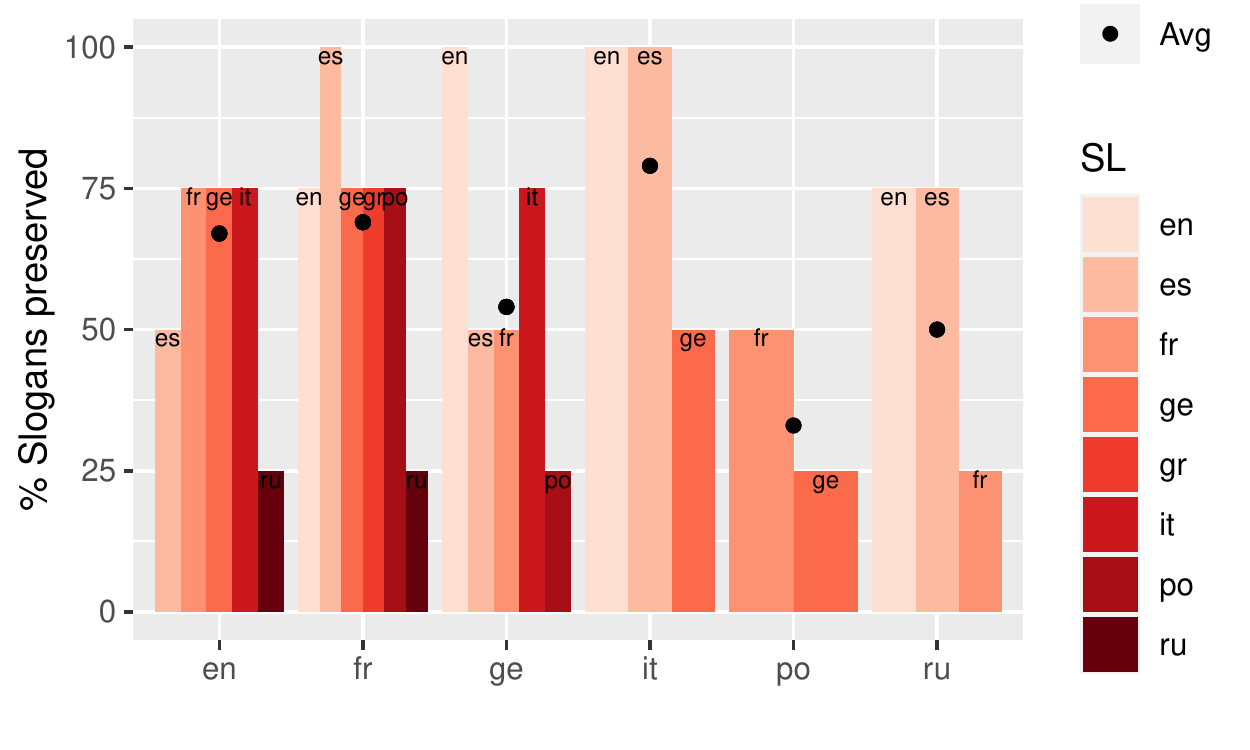}
\vspace{-0.5cm}
\captionsetup{font=footnotesize}
\caption{\% of \textit{back-translations} where {\small \textit{Slogans}} is \textit{preserved} irresp. of ({\small \textbf{Avg}}) and according to the {\small \textbf{SL}}, resp.}\label{fig:BT_slogans_preserved}
\endminipage
%\vspace{-0.3cm}
\end{figure*}

\subsection{Evaluating Augmented Data}
\label{subsec:augmented_data_eval}
%Besides the model's performance (Section X), we evaluate the quality of our data augmentation, both in an automatic and human setup.
\paragraph{Automatic Evaluation}
%backtranslations = BLEU
Common metrics to evaluate automatic translations include {\small BLEU} \cite{Papineni-et-al:2002}, {\small ROUGE} \cite{Lin-Hovy:2003}, {\small METEOR} \cite{Banerjee-Lavie:2005} and require a reference \textit{gold} (human) translation to be computed. 
We compute the {\small BLEU} scores for the paraphrases obtained via back-translations. Results vary greatly depending on the pivot language (e.g., en2\textit{ru}2en vs. en2\textit{es}2en) but on average, the scores are reasonable across languages (Tab. \ref{tab:avg_bleu}). Table \ref{tab:bleu} in App. \ref{app:data_aug} provides the detailed {\small BLEU} scores across all combinations. 
{\small BLEU} does not account, however, for fluency nor the persuasion technique preservation. Moreover, we cannot use it to evaluate our automatically obtained translations as we do not have gold references.   
We therefore conduct a small-scale human annotation study to assess the quality of our (back-)translated data.

\paragraph{Human Evaluation}
We perform human evaluation for \textit{en}, \textit{fr}, \textit{it}, \textit{ru} and \textit{ge}. %\footnote{We intend to obtain human evaluation for \textit{po} for the final submission.}. 
For \textbf{back-translations}, which we consider \textit{paraphrases} of the original input, we ask one ((near-)native) volunteer per language to rate the quality of the generated paraphrases on a scale 1--5 for the following three aspects: \textit{fluency}, \textit{fidelity}, and \textit{surface variability}, where 5 is the best score. For \textbf{translations}, as there is no reference input, the same annotators are asked to rate the quality of the generated input regarding \textit{fluency} (1-5), and whether they think this is a \textit{human-produced} paragraph. Additionally, in both setups, we ask our annotators if they think the original assigned label is \textit{preserved} (for paraphrases) or \textit{fits} in the generated paragraph (for translations). See App. \ref{app:human} for a detailed explanation of data selection and annotation setup.
\paragraph{Findings}
% this is from the appendix
%\paragraph{Analyses}
%Fig. \ref{appfig:T_fluency} and \ref{appfig:BT_fluency} present the fluency scores attributed to translations and back-translations, respectively. For each target language, we report the average scores irrespective of the source/pivot language (\textbf{Avg}) and the average scores depending on which language it was (back-)translated from. Overall, fluency scores are rather high (avg. 4), which means that (back-)translation does not affect to a large extent the readability of the generated output. However, the percentage of instances that are judged "human-produced" (Fig. \ref{appfig:T_human_produced}) drops with regards to \textit{translations} in \textit{fr, ge} and \textit{ru} (around 30\%). When zooming into language pairs, this percentages drops under 25\% for \textit{en2fr}, \textit{it2fr}, \textit{it2ge} and \textit{{en,fr}2ru}. 
%We also collected ratings regarding the \textit{fidelity} and the \textit{surface variability} aspects for the paraphrases obtained via back-translations. On average, and across target languages and language pairs, fidelity scores are found between 3 and 4, which confirms that information gets somewhat lost in back-translation. Regarding surface variability, scores do not go over 3.5: paraphrases do not diverge too much from their respective original inputs but are also not complete copy pastes, which is the desired outcome when paraphrasing. 
Due to space restrictions, we provide a complete analysis of human judgements in App. \ref{app:human} and address in this section the question: to which extent can persuasion techniques be \textit{transferred} or \textit{preserved} in the process of \textit{(back-)translation}, respectively? 
%We further address the question to which extent persuasion techniques are potentially \textit{transferred} or \textit{preserved} in the process of \textit{(back-)translation}, respectively? 
As shown in Fig. \ref{fig:T_loaded_preserved}, \textit{Loaded Language} easily transfers from \textit{all} languages to \textit{en} and \textit{fr} (avg. 85\% and 100\%, resp.), but does not when its instances are translated into \textit{ge} (avg. 52\%). The opposite is however found in back-translation, as the label is mostly preserved in \textit{ge} but not in \textit{en} and \textit{fr} (Fig. \ref{fig:BT_loaded_preserved}).
Across target languages, Fig. \ref{fig:T_slogans_preserved} and \ref{fig:BT_slogans_preserved} show that \textit{Slogans} seems more impacted than \textit{Loaded Language} by the source language from which it gets (back-)translated; 
however, this is not consistent per language across translation and back-translation. %, e.g., 
Indeed, a label can be 100\% transferable from \textit{po} to \textit{ge} but gets lost the other way around. %if attributed in an original \textit{ge} instance which gets back-translated from \textit{po}, it gets lost. 
These findings give evidence that some persuasion techniques might be language- and culture-dependent, as their \textit{transfer} and \textit{preservation} vary depending on the \textit{language pair} in consideration. Overall, the percentage of "lost labels" through (back-)translation, according to our human judgements, remains more or less low depending on the persuasion technique, indicating that they \textit{do not completely} "get lost in (back-)translations". As a result, while we expect the detection of certain persuasion techniques to benefit from our augmented data, it would not be surprising if the detection of others gets hindered by it. We present the results of our regression analysis in Sec. \ref{sec:results}. 

%However, for the translation setup, we had also included six \textit{control} original instances, i.e., in the original language, therefore not translations, to assure that potential label "loss" between all other {\small SL} to {\small TL} was not an artifact of the human judgements. It appears that identifying persuasion techniques, even when provided with a definition and examples, is a difficult task, even for humans: even though these techniques were present in the \textit{gold data}, our annotators judged that \textit{Loaded Language} was "lost" in over 70\% in \textit{ge} of the cases and that \textit{Slogans} was 100\% "lost" in \textit{en} and half the time in \textit{fr} and \textit{ge}.
%This raises the limitations of our small-scale annotation study to evaluate the quality of our automatically-obtained (back-)translations. We pave the way for an interesting project regarding the transfer and the preservation of persuasion techniques through (back-)translation, and a larger-scale human evaluation study could be conducted to confirm our findings.

\section{Our System: Approach and Methods}
\label{sec:system}

Our approach combines predictions of several models in an ensemble, which differ in three main aspects: a) training data b) model architecture and c) input format to the model. 
We show in Table \ref{tab:size-train-data} the size of each training data we used for the final task submission, and report in Table \ref{apptab:size-train-data} (App. \ref{app:data-size}) all the training data we experimented with. %of the possible training sets including the gold training data and different augmentations. 
In the following, used model architectures are presented.
%\footnote{We list approaches and models we experimented with but did not include in the final ensemble for performance reasons in App.~\ref{subsec:neg_insights}.}

\begin{table*}[!htpb]
\centering
\small
\refstepcounter{table}
\label{tab:my-table}
\begin{tabular}{lrr|c|lllc} 
\midrule
 & \multicolumn{3}{l}{Official Results ($F1$)}  & \multicolumn{4}{c}{Models + resp. training dataset(s) $\in$ Ensemble}\\ 
\midrule
     & \multicolumn{1}{l}{Micro}    & \multicolumn{1}{l}{Macro} &    \multicolumn{1}{l}{Rank}     & \multicolumn{1}{l}{{\small XLM-R-base}} & \multicolumn{1}{l}{{\small XLM-R-large}} & \multicolumn{1}{l}{Adapters} & \multicolumn{1}{l}{Additional}  \\ 
\midrule
en   & 0.26                         & 0.08                      & 15/23 & {\small +BT-sl}, {\small +T+BT}                 & {\small +BT-sl}, {\small +T+BT-sl}, {\small spans}       & {\small +T+BT-sl}                   & setfit, heur.                   \\
fr   & \textbf{0.47}                         & \textbf{0.33}                      & \textbf{1/20}  & {\small +T+BT}                 & {\small +BT-sl},  {\small spans}       & {\small +T+BT-sl}                     & -                               \\
it   & 0.54                         & 0.27                      & 2/20  &  {\small +T+BT}                  & {\small +BT-sl},  {\small spans}       & {\small +T+BT-sl}                     & -                               \\
ru   & 0.31                         & 0.19                      & 8/19  & {\small +T+BT}                  & {\small +BT-sl}                & {\small +T+BT-sl}                    & -                               \\
ge   & 0.51                         & 0.27                      & 2/20  & {\small +T+BT}                  & {\small +BT-sl},  {\small spans}       & {\small +T+BT-sl}                     & -                               \\
po   & 0.42                         & 0.25                      & 2/20  &  {\small +T+BT}                 & {\small +BT-sl},  {\small spans}       & {\small +T+BT-sl}                    & -                               \\
es   & 0.37                         & 0.18                      & 3/17  & {\small +T+BT}                  & {\small +BT-sl}, {\small +T+BT-sl},  {\small spans}       & {\small +T+BT-sl}                     & -                               \\
el   & 0.26                         & 0.16                      & 3/16  & {\small +T+BT}                  & {\small +BT-sl}, {\small +T+BT-sl},  {\small spans}       & {\small +T+BT-sl}                     & -                               \\
ka   & 0.41                         & 0.31                      & 3/16  & {\small +T+BT}                  & {\small +BT-sl}, {\small +T+BT-sl},  {\small spans}       & {\small +T+BT-sl}   
\end{tabular}
\captionsetup{font=footnotesize}
\captionsetup{width=.90\textwidth} 
\caption{Official test results and corresponding leaderboard rankings based on the official metrics {\small $F1$} micro. Note that for each test language we experiment with different possible model combinations in an ensemble and pick a different combination, depending on which ensemble results in the the highest F1-micro score on the validation set. We report {\small $F1$} macro for completeness and show which models and their respective training data were considered in the ensemble for a given language. For example, for \textit{fr} we find the best results with combining predictions of 4 different models: {\small XLM-R-base} trained with {\small +T+BT}, {\small XLM-R-large}, one trained with {\small +BT-sl} and one trained with {\small spans} and predictions by the {\small adapters} trained on {\small +T+BT-sl}.}  
\label{tab:official_results}
%\vspace{-0.1cm}
\end{table*}

\subsection{Model Architectures}
\paragraph{XLM-RoBERTa-base/large} We fine-tune all parameters of the {\small XLM-R}o{\small BERT}a ({\small XLM-R}) models with a multi-label classification head on top.

\paragraph{Adapter} We train a label-specific adapter for each persuasion technique. Adapters \cite{Houlsby2019ParameterEfficientTL} are a specific set of parameters inserted in every layer of a transformer. Instead of fine-tuning the parameters of the full pre-trained language model, these smaller parameters are updated for a specific task while the rest of the parameters is kept frozen. This makes them more efficient to train while still being compatible with the original transformer architecture. We use {\small XLM-R-base} as a back-bone model and the binary cross-entropy loss for each label. After training, we combine predictions of the 23 adapters for each paragraph. Note that the adapters are especially useful for low-frequency classes as the binary classification setup usually leads to a higher recall for such labels. 

\paragraph{SetFit} %SetFit \cite{setfit} is a few-shot learning method that is based on sentence-transformers. 
This few-shot learning method is based on sentence-transformers \cite{setfit}. As a first step, a pre-trained {\small SBERT} model is fine-tuned on a small number of labeled text pairs in a contrastive Siamese manner. This model can then be used to generate embeddings for sentences or paragraphs and to train a simple text classifier for the target task. The main advantages compared to other few-shot fine-tuning approaches are (i) efficiency and (ii) that it does not require prompts or verbalizers. Using a multi-target strategy, we train a distinct logistic regression classifier for each persuasion strategy with paragraph representations as inputs, which are then combined to output a prediction for each label for each paragraph. 

\subsection{Training data}
Our data augmentation techniques (Sec. \ref{sec:data}) allow us to train our models on different training sets of different sizes, which we report in Table \ref{tab:size-train-data}.
%As described in Section \ref{sec:data}, we use data augmentation to create different sets of training data to train our models on. 
Besides the original \textit{gold} training data, we obtain six additional training corpora as a result of \textit{translations} ({\small T}) and \textit{back-translations} ({\small BT}) techniques. 
Additionally, we experiment with injecting in the \textit{gold} training data the relevant textual \textit{spans} triggering the annotated persuasion techniques, 
%Besides the original training data (\textit{gold}) and the different augmented variants we add another variant, \textit{span}, for which we add the corresponding spans in addition to the full paragraphs of gold training data.
% TO-DO: another sentence how the spans are extracted / added. Prisca: I added this info in the data section, so that we don't have to repeat it here.
%This should help 
in the hope that it helps the models to particularly focus on relevant information for each persuasion technique.
% Did we train on only spans and it was worse?
% Pri: I did, but with robertabase, and yes, it sucked. it's reported as XLM-spans in the protocole, (it was only on dev set)
% I only provided results for the usual six languages + es (why?)
%Tab.~ \ref{tab:size-train-data} shows an overview of all possible training corpora.

\subsection{Post-processing and ensemble}
After training each model we apply \textbf{threshold moving}, i.e., given the validation set for each language, we search the optimal classification threshold for each model. We tune the threshold for a range between 0.1 and 0.9 (step size$=$0.1) and pick the one that maximizes the {\small F1}-micro score on the corresponding validation set.
We also develop rule-based \textbf{language-specific heuristics} for a small number of labels. For instance, for the persuasion technique \textit{Doubt}, we overturn the model predictions iff a paragraph contains, for example, a question mark or question words.

We then compute the {\small F1}-scores on the validation set using an \textbf{ensemble}. For each instance in the validation set, all models that are part of the ensemble can vote (predict the classes according to the corresponding threshold). 
\textbf{
\begin{table}[!htpb]
\centering
\small
\begin{tabular}{@{}llrl@{}}
\toprule
                                  & \textbf{Df} & \textbf{explvar} & \textbf{sign} \\ \midrule
\textbf{trainingSet:label}        & 110         & 38.00            & ***           \\
\textbf{label}                    & 22          & 31.31            & ***           \\
\textbf{trainingSet}              & 5           & 18.28            & ***           \\
\textbf{testLang:trainingSet}     & 25          & 0.89             & ***           \\
\textbf{testLang}                 & 5           & 0.24             & ***           \\
\textbf{total explained variance} &             & \textbf{88.73}   &               \\ \bottomrule
\end{tabular}
\captionsetup{font=footnotesize}
\caption{Terms of the most explanatory regression model for predicting \textbf{F1 (persuasion strategy)}, with degrees of freedom, statistical significance and explained variance. The best fit explains 88.73\% of the variance.}
\label{regressionresult}
%\vspace{-0.1cm}
\end{table}}
For each class, the votes are then summed up. For the final prediction, a class is added if the sum of votes exceeds the voting threshold. 
We calculate the optimal voting threshold again based on the validation set {\small F1}-micro score.

\subsection{Experimental Setup}
\label{sec:experimental}

We train {\small XLM-R-base} for 10 epochs on the different training datasets and apply early stopping, picking the model that achieves the highest {\small F1}-micro score on the provided development dataset. 
We do the same with {\small XLM-R-large} with a maximum of 5 epochs. The adapters are trained only on the largest training dataset ({\small +T+BT-sl}) for a maximum of 5 epochs. SetFit is trained on 1,000 instances sampled from {\small +T+BT-sl}. More details %about the hyper-parameter setup and 
on implementation can be found in Table \ref{tab:hyperparameter} in App. \ref{subsec:models}\footnote{Note that for \textit{ru}, the \textit{span} model was not included in the final ensemble due to a model error.}. 
We also report our non-submitted experiments in App. \ref{subsec:neg_insights}, as we draw some insightful conclusions. %\footnote{All our experiments, included the non-submitted, are listed in App. \ref{subsec:neg_insights}. Note that for \textit{ru}, the \textit{span} model was not included in the final ensemble due to a model error.}

\section{Results and Analysis}
\label{sec:results}

Subtask 3 official %test 
results and corresponding leaderboard rankings are shown in Tab.~\ref{tab:official_results}.
Note that for each language we find an ensemble by computing the F1-micro score for different model combinations. The combination for each test language is listed in Tab.~\ref{tab:official_results}, including the respective training data.
We win the task for \textit{fr}, achieve 2\ts{nd} places for \textit{po}, \textit{it}, \textit{ge} and  3\ts{rd} places for the surprise languages \textit{el}, \textit{gr}, \textit{ka}. We rank roughly around mid-field for \textit{ru} and \textit{en}, but manage a 3\ts{rd} place post-submission for the latter. 

%\paragraph{Quantitative Analysis}: 
To identify trends in the effectiveness of the various data augmentation strategies employed, we 
train a linear regression model to predict the {\small F1} 
\begin{figure}[!ht]
    \centering
    \includegraphics[width=8cm,height=7cm]{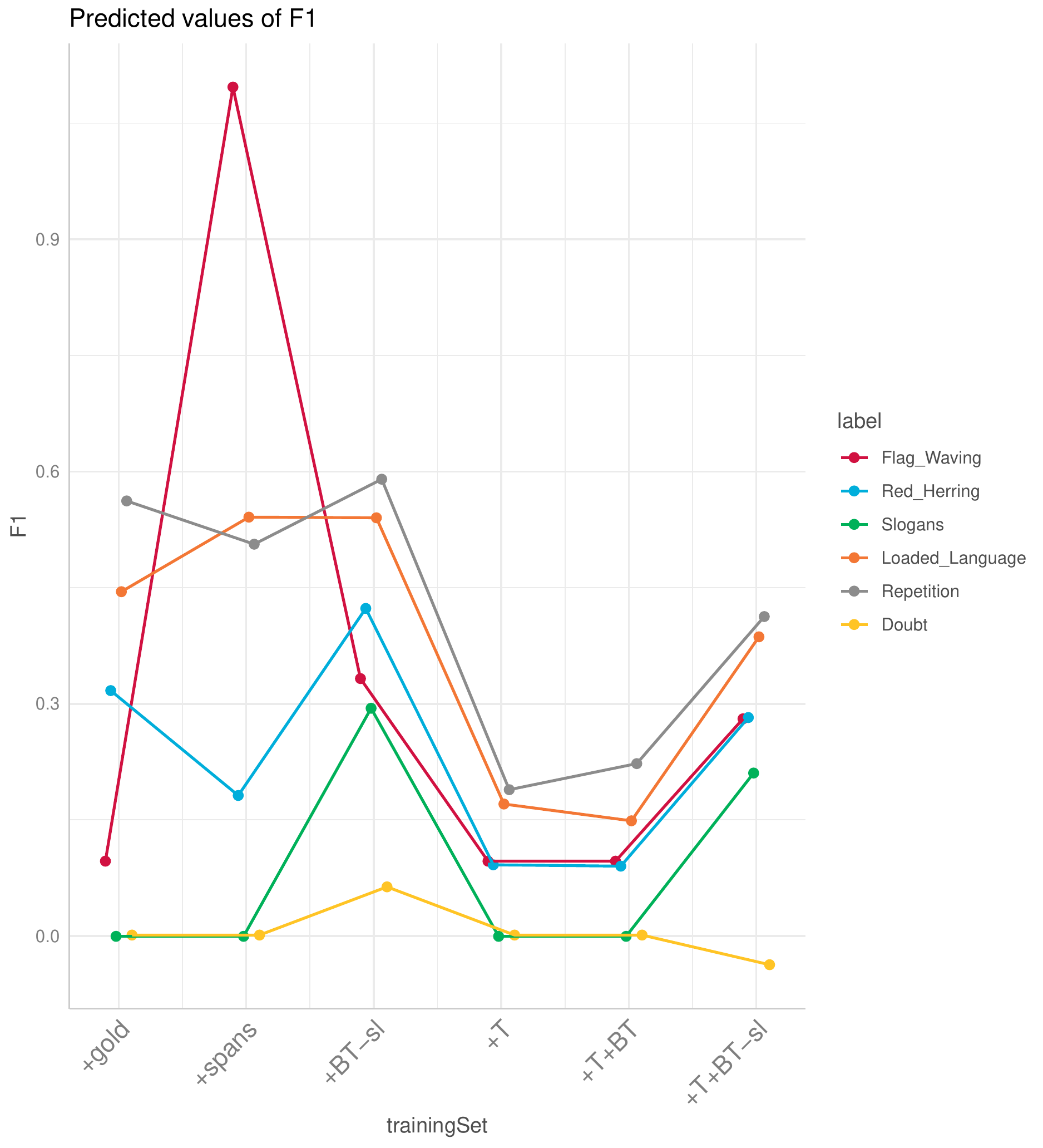}
    %\vspace{-0.3cm}
    \captionsetup{font=footnotesize}
    \caption{Effect of training data type (+size, in ascending order left to right) on predicted {\small F1} scores for 6 persuasion techniques.}
    %\vspace{-0.5cm}
    \label{interaction_corpus_label}
\end{figure}scores of the different persuasion labels as dependent variables. 
Our regression model looks at the effect of \textit{training data}, \textit{test language} and \textit{persuasion technique}, each coded as categorical independent variables ({\small IV})\footnote{App.~\ref{sec:regression} presents all details of the modeling set up.}.
%To address the questions which factors play the most important role in predicting different persuasion strategies effectively
Regression results are presented in Tab.~\ref{regressionresult}. We first look at which factors play an important role in predicting different persuasion strategies effectively. 
We look at the amount of explained variance by each {\small IV}: the best fit ($Adj. R^2 = 0.86$) contains each {\small IV} and the interactions between each training data with (i) test languages and (ii) persuasion techniques. 
Very little systematic differences across test languages are observed, as there is 1\% explained variance by the interaction between training data and test languages.
In contrast, most variance is explained by the interaction between training data and labels (38\%) indicating that %the choice of the augmentation strategy plays an important role with varying effects on performance w.r.t persuasion techniques. 
performance in detecting persuasion techniques is considerably impacted by the choice of  augmentation strategy. %\footnote{More details in Tab.~\ref{regressionresult}, App.~\ref{sec:regression}}.

%In contrast, there are little systematic differences across the test languages ($1\%$ explained variance by interaction between training set and test lang). 
We now zoom into the extent to which the choice of augmentation strategy differs across persuasion techniques detection. 
We visualize the interaction term as an effect plot in Fig. \ref{interaction_corpus_label}. We select six labels of interest to show the overall trend, and present the full plots in App.~\ref{sec:regression}. %\footnote{We selected 6 labels of interest to show the overall trend. The full plot is in the appendix (REF FULL PLOT)}. 
%The x-axis shows the different augmented training sets sorted by size. 
We find the different augmented training sets on the x-axis (sorted by ascending size) and their impact on {\small F1} scores' predictions on the y-axis. 
We observe %the following trends:
that in general, data augmentation positively impacts the predicted performance, especially for the less-frequent persuasion techniques, e.g., \textit{Flag Waving}. %\textit{Doubt, Flag Waving, Red Herring}.
Training on the original \textit{gold} data with the relevant textual \textit{spans} can have extremely positive effects (\textit{Flag Waving}, \textit{Loaded Language}) %has extremely positive effects on \textit{Flag Waving} (in red) and slightly positive effects on \textit{Loaded Language} (orange). However, 
but this effect is not observed across \textit{all} labels, indicating that some persuasion techniques' relevant textual information is particularly compressed and the context is therefore less crucial. 
Results clearly point to \textit{back-translation} ({\small +BT-sl}) as the most robust augmentation strategy with consistent improvement of {\small F1} scores across all labels.

%However, augmented data %augmenting the training data %via augmentation 
%is not always beneficial: adding 
In contrast, \textit{translations} consistently hurt the performance across all labels and is only effective if combined with \textit{back-translations} (e.g., {\small +T+BT-sl}).
%higher F1 for \textit{Slogans} if trained on +T+BT-sl).
This considerable difference in performance between injecting translations vs. back-translations, which are inherently the same processes, is surprising and not \textit{on par} with our human evaluation findings (Sec~\ref{subsec:augmented_data_eval}); we plan to conduct further analyses to investigate the phenomenon.  
Overall, these analyses have shown that adding \textit{some more} data, i.e., {\small +BT} only, does indeed improve performance but \textit{too much} augmented data, i.e., {\small +T, +T-BT} tends to hinder it.

\section{Conclusion}
\label{sec:conclusion}
We tackled the task of detecting persuasion techniques in online news in a multi-lingual setup. %by
We built a system that successfully combines natural with augmented data via (back-)translation %data augmentation techniques via (back-)translation 
with an ensemble of {\small SOTA} multi-lingual models. 
While we showed that using augmented data, i.e., \textit{more} data, generally boosts performance, our results also indicate that %performance on detecting frequent persuasion techniques 
it might be hurt when integrating \textit{too much} augmented data. %in training.
% balancing gold and augmented data is key.
In conclusion, for the task of persuasion techniques detection, \textit{more} data, obtained via (back-)translation, does help overall, but \textit{less might be more} when it comes to adding automatically-generated translations.%: balancing natural vs. augmented data is key. 
%\newpage
\section*{Ethics Statement} \label{sec:ethics}
In the context of our evaluation task, we collected ratings from human participants. For this, the participants were provided an Informed Consent Letter with the name and the contact of the investigators; the title, purpose and procedure of the study; risks and benefits for participating in the study; confirmation of confidential anonymous data handling; and confirmation that participation in the study is voluntary. The Informed Consent Letter was signed before the participants took part in the study.
\section*{Acknowledgements}
\label{sec:acknowledgements}
We thank the volunteering annotators taking part in the annotation study. We are grateful for helpful feedback from Gabriella Lapesa and the anonymous reviewers.

Neele Falk is supported by Bundesministerium für Bildung und Forschung (BMBF) through the project E-DELIB (Powering up e-deliberation: towards AI-supported moderation).
Annerose Eichel receives funding by the Hanns-Seidel-Stiftung. Prisca Piccirilli is supported by the Studienstiftung des deutschen Volkes and the DFG Research Grant SCHU 2580/4-1 \textit{Multimodal Dimensions and Computational Applications of Abstractness}. 

%annotators + whoever gave feedback like Gabriella 
%\input{sections/limitations}
%\input{sections/ethics_statement}

\bibliography{custom}

\begin{thebibliography}{21}
\expandafter\ifx\csname natexlab\endcsname\relax\def\natexlab#1{#1}\fi

\bibitem[{Abujaber et~al.(2021)Abujaber, Qarqaz, and
  Abdullah}]{Abujaber-et-al:2021}
Dia Abujaber, Ahmed Qarqaz, and Malak~A. Abdullah. 2021.
\newblock \href {https://aclanthology.org/2021.semeval-1.148} {{L}e{C}un at
  {S}em{E}val-2021 task 6: Detecting persuasion techniques in text using
  ensembled pretrained transformers and data augmentation}.
\newblock In \emph{Proceedings of the 15th International Workshop on Semantic
  Evaluation (SemEval-2021)}, pages 1068--1074, Online. Association for
  Computational Linguistics.

\bibitem[{Banerjee and Lavie(2005)}]{Banerjee-Lavie:2005}
Satanjeev Banerjee and Alon Lavie. 2005.
\newblock \href {https://aclanthology.org/W05-0909} {{METEOR}: An automatic
  metric for {MT} evaluation with improved correlation with human judgments}.
\newblock In \emph{Proceedings of the {ACL} Workshop on Intrinsic and Extrinsic
  Evaluation Measures for Machine Translation and/or Summarization}, pages
  65--72, Ann Arbor, Michigan. Association for Computational Linguistics.

\bibitem[{Chernyavskiy et~al.(2020)Chernyavskiy, Ilvovsky, and
  Nakov}]{chernyavskiy-etal-2020-aschern}
Anton Chernyavskiy, Dmitry Ilvovsky, and Preslav Nakov. 2020.
\newblock \href {https://doi.org/10.18653/v1/2020.semeval-1.191} {Aschern at
  {S}em{E}val-2020 task 11: It takes three to tango: {R}o{BERT}a, {CRF}, and
  transfer learning}.
\newblock In \emph{Proceedings of the Fourteenth Workshop on Semantic
  Evaluation}, pages 1462--1468, Barcelona (online). International Committee
  for Computational Linguistics.

\bibitem[{Da~San~Martino et~al.(2019)Da~San~Martino, Barr{\'o}n-Cede{\~n}o, and
  Nakov}]{da-san-martino-etal-2019-findings}
Giovanni Da~San~Martino, Alberto Barr{\'o}n-Cede{\~n}o, and Preslav Nakov.
  2019.
\newblock \href {https://doi.org/10.18653/v1/D19-5024} {Findings of the
  {NLP}4{IF}-2019 shared task on fine-grained propaganda detection}.
\newblock In \emph{Proceedings of the Second Workshop on Natural Language
  Processing for Internet Freedom: Censorship, Disinformation, and Propaganda},
  pages 162--170, Hong Kong, China. Association for Computational Linguistics.

\bibitem[{Da~San~Martino et~al.(2020)Da~San~Martino, Barr{\'o}n-Cede{\~n}o,
  Wachsmuth, Petrov, and Nakov}]{da-san-martino-etal-2020-semeval}
Giovanni Da~San~Martino, Alberto Barr{\'o}n-Cede{\~n}o, Henning Wachsmuth,
  Rostislav Petrov, and Preslav Nakov. 2020.
\newblock \href {https://doi.org/10.18653/v1/2020.semeval-1.186}
  {{S}em{E}val-2020 task 11: Detection of propaganda techniques in news
  articles}.
\newblock In \emph{Proceedings of the Fourteenth Workshop on Semantic
  Evaluation}, pages 1377--1414, Barcelona (online). International Committee
  for Computational Linguistics.

\bibitem[{Dimitrov et~al.(2021)Dimitrov, Bin~Ali, Shaar, Alam, Silvestri,
  Firooz, Nakov, and Da~San~Martino}]{dimitrov-etal-2021-semeval}
Dimitar Dimitrov, Bishr Bin~Ali, Shaden Shaar, Firoj Alam, Fabrizio Silvestri,
  Hamed Firooz, Preslav Nakov, and Giovanni Da~San~Martino. 2021.
\newblock \href {https://doi.org/10.18653/v1/2021.semeval-1.7}
  {{S}em{E}val-2021 task 6: Detection of persuasion techniques in texts and
  images}.
\newblock In \emph{Proceedings of the 15th International Workshop on Semantic
  Evaluation (SemEval-2021)}, pages 70--98, Online. Association for
  Computational Linguistics.

\bibitem[{Houlsby et~al.(2019)Houlsby, Giurgiu, Jastrzebski, Morrone,
  De~Laroussilhe, Gesmundo, Attariyan, and
  Gelly}]{Houlsby2019ParameterEfficientTL}
Neil Houlsby, Andrei Giurgiu, Stanislaw Jastrzebski, Bruna Morrone, Quentin
  De~Laroussilhe, Andrea Gesmundo, Mona Attariyan, and Sylvain Gelly. 2019.
\newblock \href {https://proceedings.mlr.press/v97/houlsby19a.html}
  {Parameter-efficient transfer learning for {NLP}}.
\newblock In \emph{Proceedings of the 36th International Conference on Machine
  Learning}, volume~97 of \emph{Proceedings of Machine Learning Research},
  pages 2790--2799. PMLR.

\bibitem[{Junczys-Dowmunt et~al.(2018{\natexlab{a}})Junczys-Dowmunt,
  Grundkiewicz, Dwojak, Hoang, Heafield, Neckermann, Seide, Germann, Fikri~Aji,
  Bogoychev, Martins, and Birch}]{Junczys-et-al:2018}
Marcin Junczys-Dowmunt, Roman Grundkiewicz, Tomasz Dwojak, Hieu Hoang, Kenneth
  Heafield, Tom Neckermann, Frank Seide, Ulrich Germann, Alham Fikri~Aji,
  Nikolay Bogoychev, Andr\'{e} F.~T. Martins, and Alexandra Birch.
  2018{\natexlab{a}}.
\newblock \href {http://www.aclweb.org/anthology/P18-4020} {Marian: Fast neural
  machine translation in {C++}}.
\newblock In \emph{Proceedings of ACL 2018, System Demonstrations}, pages
  116--121, Melbourne, Australia. Association for Computational Linguistics.

\bibitem[{Junczys-Dowmunt et~al.(2018{\natexlab{b}})Junczys-Dowmunt,
  Grundkiewicz, Dwojak, Hoang, Heafield, Neckermann, Seide, Germann, Fikri~Aji,
  Bogoychev, Martins, and Birch}]{mariannmt}
Marcin Junczys-Dowmunt, Roman Grundkiewicz, Tomasz Dwojak, Hieu Hoang, Kenneth
  Heafield, Tom Neckermann, Frank Seide, Ulrich Germann, Alham Fikri~Aji,
  Nikolay Bogoychev, Andr\'{e} F.~T. Martins, and Alexandra Birch.
  2018{\natexlab{b}}.
\newblock \href {http://www.aclweb.org/anthology/P18-4020} {Marian: Fast neural
  machine translation in {C++}}.
\newblock In \emph{Proceedings of ACL 2018, System Demonstrations}, pages
  116--121, Melbourne, Australia. Association for Computational Linguistics.

\bibitem[{Jurkiewicz et~al.(2020)Jurkiewicz, Borchmann, Kosmala, and
  Grali{\'n}ski}]{jurkiewicz-etal-2020-applicaai}
Dawid Jurkiewicz, {\L}ukasz Borchmann, Izabela Kosmala, and Filip
  Grali{\'n}ski. 2020.
\newblock \href {https://doi.org/10.18653/v1/2020.semeval-1.187} {{A}pplica{AI}
  at {S}em{E}val-2020 task 11: On {R}o{BERT}a-{CRF}, span {CLS} and whether
  self-training helps them}.
\newblock In \emph{Proceedings of the Fourteenth Workshop on Semantic
  Evaluation}, pages 1415--1424, Barcelona (online). International Committee
  for Computational Linguistics.

\bibitem[{Lin and Hovy(2003)}]{Lin-Hovy:2003}
Chin-Yew Lin and Eduard Hovy. 2003.
\newblock \href {https://aclanthology.org/N03-1020} {Automatic evaluation of
  summaries using n-gram co-occurrence statistics}.
\newblock In \emph{Proceedings of the 2003 Human Language Technology Conference
  of the North {A}merican Chapter of the Association for Computational
  Linguistics}, pages 150--157.

\bibitem[{Mapes et~al.(2019)Mapes, White, Medury, and
  Dua}]{mapes-etal-2019-divisive}
Norman Mapes, Anna White, Radhika Medury, and Sumeet Dua. 2019.
\newblock \href {https://doi.org/10.18653/v1/D19-5014} {Divisive language and
  propaganda detection using multi-head attention transformers with deep
  learning {BERT}-based language models for binary classification}.
\newblock In \emph{Proceedings of the Second Workshop on Natural Language
  Processing for Internet Freedom: Censorship, Disinformation, and Propaganda},
  pages 103--106, Hong Kong, China. Association for Computational Linguistics.

\bibitem[{Papineni et~al.(2002)Papineni, Roukos, Ward, and
  Zhu}]{Papineni-et-al:2002}
Kishore Papineni, Salim Roukos, Todd Ward, and Wei-Jing Zhu. 2002.
\newblock \href {https://aclanthology.org/P02-1040} {{B}leu: a method for
  automatic evaluation of machine translation}.
\newblock In \emph{Proceedings of the 40th Annual Meeting of the Association
  for Computational Linguistics}, pages 311--318, Philadelphia, Pennsylvania,
  USA. Association for Computational Linguistics.

\bibitem[{Pedregosa et~al.(2011)Pedregosa, Varoquaux, Gramfort, Michel,
  Thirion, Grisel, Blondel, Prettenhofer, Weiss, Dubourg, Vanderplas, Passos,
  Cournapeau, Brucher, Perrot, and Duchesnay}]{scikit-learn}
F.~Pedregosa, G.~Varoquaux, A.~Gramfort, V.~Michel, B.~Thirion, O.~Grisel,
  M.~Blondel, P.~Prettenhofer, R.~Weiss, V.~Dubourg, J.~Vanderplas, A.~Passos,
  D.~Cournapeau, M.~Brucher, M.~Perrot, and E.~Duchesnay. 2011.
\newblock Scikit-learn: Machine learning in {P}ython.
\newblock \emph{Journal of Machine Learning Research}, 12:2825--2830.

\bibitem[{Pfeiffer et~al.(2020)Pfeiffer, R{\"u}ckl{\'e}, Poth, Kamath,
  Vuli{\'c}, Ruder, Cho, and Gurevych}]{pfeiffer2020AdapterHub}
Jonas Pfeiffer, Andreas R{\"u}ckl{\'e}, Clifton Poth, Aishwarya Kamath, Ivan
  Vuli{\'c}, Sebastian Ruder, Kyunghyun Cho, and Iryna Gurevych. 2020.
\newblock Adapterhub: A framework for adapting transformers.
\newblock In \emph{Proceedings of the 2020 Conference on Empirical Methods in
  Natural Language Processing: System Demonstrations}, pages 46--54.

\bibitem[{Piskorski et~al.(2023)Piskorski, Stefanovitch, Da~San~Martino, and
  Nakov}]{semeval2023task3}
Jakub Piskorski, Nicolas Stefanovitch, Giovanni Da~San~Martino, and Preslav
  Nakov. 2023.
\newblock Semeval-2023 task 3: Detecting the category, the framing, and the
  persuasion techniques in online news in a multi-lingual setup.
\newblock In \emph{Proceedings of the 17th International Workshop on Semantic
  Evaluation}, SemEval 2023, Toronto, Canada.

\bibitem[{Tian et~al.(2021)Tian, Gui, Li, Yan, and Xiao}]{tian-etal-2021-mind}
Junfeng Tian, Min Gui, Chenliang Li, Ming Yan, and Wenming Xiao. 2021.
\newblock \href {https://doi.org/10.18653/v1/2021.semeval-1.150} {{M}in{D} at
  {S}em{E}val-2021 task 6: Propaganda detection using transfer learning and
  multimodal fusion}.
\newblock In \emph{Proceedings of the 15th International Workshop on Semantic
  Evaluation (SemEval-2021)}, pages 1082--1087, Online. Association for
  Computational Linguistics.

\bibitem[{Tiedemann and Thottingal(2020)}]{Tiedemann-Thottingal:2020}
J{\"o}rg Tiedemann and Santhosh Thottingal. 2020.
\newblock {OPUS-MT} — {B}uilding open translation services for the {W}orld.
\newblock In \emph{Proceedings of the 22nd Annual Conferenec of the European
  Association for Machine Translation}, Lisbon, Portugal.

\bibitem[{Troiano et~al.(2020)Troiano, Klinger, and
  Pad{\'o}}]{Troiano-et-al:2020}
Enrica Troiano, Roman Klinger, and Sebastian Pad{\'o}. 2020.
\newblock \href {https://aclanthology.org/2020.coling-main.384} {Lost in
  back-translation: Emotion preservation in neural machine translation}.
\newblock In \emph{Proceedings of the 28th International Conference on
  Computational Linguistics}, Barcelona, Spain (Online). International
  Committee on Computational Linguistics.

\bibitem[{Tunstall et~al.(2022)Tunstall, Reimers, Jo, Bates, Korat, Wasserblat,
  and Pereg}]{setfit}
Lewis Tunstall, Nils Reimers, Unso Eun~Seo Jo, Luke Bates, Daniel Korat, Moshe
  Wasserblat, and Oren Pereg. 2022.
\newblock \href {https://doi.org/10.48550/ARXIV.2209.11055} {Efficient few-shot
  learning without prompts}.

\bibitem[{Wolf et~al.(2020)Wolf, Debut, Sanh, Chaumond, Delangue, Moi, Cistac,
  Rault, Louf, Funtowicz, Davison, Shleifer, von Platen, Ma, Jernite, Plu, Xu,
  Scao, Gugger, Drame, Lhoest, and Rush}]{wolf-etal-2020-transformers}
Thomas Wolf, Lysandre Debut, Victor Sanh, Julien Chaumond, Clement Delangue,
  Anthony Moi, Pierric Cistac, Tim Rault, Rémi Louf, Morgan Funtowicz, Joe
  Davison, Sam Shleifer, Patrick von Platen, Clara Ma, Yacine Jernite, Julien
  Plu, Canwen Xu, Teven~Le Scao, Sylvain Gugger, Mariama Drame, Quentin Lhoest,
  and Alexander~M. Rush. 2020.
\newblock \href {https://www.aclweb.org/anthology/2020.emnlp-demos.6}
  {Transformers: State-of-the-art natural language processing}.
\newblock In \emph{Proceedings of the 2020 Conference on Empirical Methods in
  Natural Language Processing: System Demonstrations}, pages 38--45, Online.
  Association for Computational Linguistics.

\end{thebibliography}

\appendix
\section{Appendix A: Data}
\label{app:a}

\subsection{Gold Label Distribution and Training Data Sizes}
\label{app:data-size}

\begin{table*}[!htpb]
\centering
%\small
\begin{tabular}{@{}lrrrrrr@{}}
\toprule
Label                             & en   & fr  & ge  & it  & po  & ru  \\ \midrule
\textbf{Justification:} & & & & & &\\
Appeal to Authority             & 154  & 76  & 225 & 70  & 41  & 10  \\
Appeal to Popularity            & 15   & 82  & 63  & 37  & 30  & 8   \\
Appeal to Values                & 0    & 100 & 73  & 131 & 101 & 48  \\
Appeal to Fear-Prejudice        & 310  & 210 & 182 & 285 & 108 & 54  \\
Flag Waving                      & 287  & 37  & 65  & 35  & 68  & 42  \\
\textbf{Simplification:} & & & & & &\\
Causal Oversimplification        & 213  & 125 & 33  & 50  & 12  & 39  \\
False Dilemma-No Choice         & 122  & 73  & 41  & 61  & 12  & 28  \\
Consequential Oversimplification & 0    & 112 & 35  & 29  & 24  & 70  \\
\textbf{Distraction:} & & & & & &\\
Straw Man                        & 15   & 135 & 15  & 51  & 15  & 21  \\
Whataboutism                      & 16   & 62  & 13  & 8   & 8   & 7   \\
Red Herring                      & 44   & 55  & 30  & 23  & 12  & 2   \\
\textbf{Call:} & & & & & &\\
Appeal to Time                  & 0    & 41  & 11  & 27  & 14  & 28  \\
Slogans                           & 153  & 149 & 87  & 54  & 36  & 72  \\
Conversation Killer              & 91   & 170 & 121 & 178 & 50  & 88  \\
\textbf{Manipulative Wording:}& & & & & &\\
Loaded Language                  & 1,809 & 944 & 242 & 903 & 310 & 641 \\
Repetition                        & 544  & 92  & 8   & 22  & 13  & 69  \\
Exaggeration-Minimisation         & 466  & 258 & 157 & 143 & 111 & 131 \\
Obfuscation-Vagueness-Confusion   & 18   & 113 & 62  & 21  & 36  & 19  \\
\textbf{Attack to Reputation:}& & & & & &\\
Appeal to Hypocrisy             & 40   & 134 & 136 & 82  & 162 & 103 \\
Doubt                             & 518  & 327 & 288 & 882 & 295 & 509 \\
Name Calling-Labeling            & 979  & 428 & 734 & 566 & 475 & 253 \\
Guilt by Association            & 59   & 130 & 122 & 53  & 94  & 24  \\
Questioning the Reputation      & 0    & 348 & 310 & 383 & 164 & 303 \\ \bottomrule
\end{tabular}
\caption{Label distributions of \textit{gold} training data (in absolute numbers), divided by coarse-grained categories. }
\label{tab:label_distribution}
\end{table*}

\begin{table*}[!htpb]
%\small
\centering
\begin{tabular}{@{} l*{7}r @{}}
\toprule
\multicolumn{6}{r}{Training Datasets}
\\\cmidrule{2-8}
 & gold & +T & +BT & +BT-sl & +T+BT & +T+BT-sl & +span  \\\midrule
 en & 3,761 & 10,928 & 18,801 & 22,561 & 25,968 & 29,728 & 7,521\\
 fr & 1,694 & 10,928 & 8,466 & 11,852 & 17,700 & 21,086 & 3,387\\
 it & 1,746 & 6,758 & 5,236 & 6,981 & 10,248 & 11,993 & 3,491\\
 ru & 1,246 & 6,699 & 3,736 & 4,981 & 9,189 & 10,434 & 2,491 \\
 ge & 1,253 & 9,683 & 6,261 & 7,513 & 14,691 & 15,943 & 2,505\\
 po & 1,233 & 4,178 & 3,697 & 3,697 & 6,642 & 6,642 & 2,465\\
 es & 0 & 9,696 & 0 & 0 & 0 & 0 & 0\\
 el & 0 & 6,706 & 0 & 0 & 0 & 0 & 0\\
 ka & 0 & 0 & 0 & 0 & 0 & 0 & 0\\
 \textbf{total} & \textbf{10,933} & \textbf{65,576} & \textbf{46,197} & \textbf{57,585} & \textbf{84,438} & \textbf{95,826} &  \textbf{21,860} \\\bottomrule
\end{tabular}
\captionsetup{width=.80\textwidth}
\caption{Training data size per language. \textit{gold} is the original task data, to which are added all possible (back-)translations (+{\small T} and +{\small BT}) - with or without the surprise languages (\textit{sl}) as pivot languages - and the relevant textual \textit{span}s.}
\label{apptab:size-train-data}
\vspace{-0.3cm}
\end{table*}

Tab.~ \ref{tab:label_distribution} lists the number of instances per label for each language for the \textit{gold} training corpus as provided by the task organizers. The overview shows that there is quite some imbalance, e.g. for each \textit{en} we observe a  maximum label frequency of $>$1,800 instances for the label \textit{Loaded Language}, while there is no data at all for the four labels \textit{Appeal to Time, Appeal to Values, Consequential Oversimplification, Questioning the Reputation}. This picture changes depending on the language.

Tab.~ \ref{apptab:size-train-data} presents an overview of all training data and their sizes we experiment with. \textit{gold} is the original data provided by the organizers, which we augment with our automatically obtained (back-)translations and the relevant persuasion technique textual \textit{spans} (provided for the task). We obtain (back-)translations from and to all possible six languages (\textit{en, it, fr, ge, ru, po}). When the organizers release the three surprise languages (\textit{es, el, ka}), we are able to obtain translations in \textit{es} and \textit{el}. We distinguish the back-translated augmented data containing - or not - the back-translations in the original six languages \textit{from} the surprise languages. However, note that when combining \textit{translations} and \textit{back-translations} (+{\small T}+{\small BT}(-sl)), we do not include the \textit{surprise language translations} (size 0). Overall, our data augmentation techniques allow us to considerably increase our training data by almost 900\%. 

\subsection{Augmented Data: Automatic Evaluation} \label{app:data_aug}

Tab.~\ref{tab:bleu} presents {\small BLEU} scores for paraphrases that were obtained via back-translation for 1,2,3,4-grams. Scores are presented per language and language direction.
Scores are rather low for back-translations in \textit{fr} via \textit{po} (11.19) and in \textit{ru} via \textit{fr} (15.62) but are overall reasonable across language pairs, reaching around 50 in \textit{it} and \textit{fr} and up to 60 for back-translation in \textit{en} via \textit{es}, giving us the intuition that the quality of these paraphrases are rather good. We conduct human evaluation to confirm this hypothesis. 

\begin{table}[!ht]
%\vspace{-0.3cm}
\small
\begin{tabular}{lrrrr}
\toprule
\textbf{lang pair} &  \textbf{1-gram} &  \textbf{2-gram} &  \textbf{3-gram} &  \textbf{4-gram} \\
\midrule
     \textbf{en2ru2en} &         54.83 &        43.25 &         35.31 &       28.76 \\
     \textbf{en2es2en} &         80.37 &        72.44 &         66.30 &       60.63 \\
     \textbf{en2it2en} &         73.45 &        64.84 &         58.27 &       52.31 \\
     \textbf{en2fr2en} &         68.01 &        59.47 &         53.03 &       47.23 \\
     \textbf{en2ge2en} &         73.13 &        64.42 &         57.61 &       51.39 \\ \midrule
     \textbf{fr2ge2fr} &         60.33 &        50.37 &         43.26 &       37.04 \\
     \textbf{fr2es2fr} &         67.23 &        59.52 &         53.63 &       48.16 \\
     \textbf{fr2po2fr} &         20.98 &        16.47 &         13.57 &       11.19 \\
     \textbf{fr2en2fr} &         69.32 &        61.22 &         55.04 &       49.33 \\
     \textbf{fr2el2fr} &         48.15 &        40.03 &         34.26 &       29.21 \\
     \textbf{fr2ru2fr} &         46.08 &        34.82 &         27.67 &       21.89 \\ \midrule
     \textbf{it2ge2it} &         55.55 &        45.12 &         37.96 &       31.86 \\
     \textbf{it2es2it} &         71.05 &        63.80 &         58.23 &       52.99 \\
     \textbf{it2en2it} &         73.52 &        64.90 &         58.44 &       52.52 \\ \midrule
     \textbf{ru2es2ru} &         48.52 &        35.68 &         27.68 &       21.39 \\
     \textbf{ru2fr2ru} &         41.69 &        28.80 &         21.25 &       15.62 \\
     \textbf{ru2en2ru} &         52.04 &        39.99 &         32.03 &       25.52 \\ \midrule
     \textbf{po2fr2po} &         40.33 &        30.60 &         24.29 &       19.23 \\
     \textbf{po2ge2po} &         48.24 &        36.70 &         29.25 &       23.22 \\ \midrule
     \textbf{ge2it2ge} &         56.47 &        44.25 &         36.33 &       29.86 \\
     \textbf{ge2po2ge} &         53.43 &        40.49 &         32.35 &       25.89 \\
     \textbf{ge2es2ge} &         61.21 &        48.51 &         40.06 &       33.13 \\
     \textbf{ge2fr2ge} &         57.65 &        45.59 &         37.57 &       31.04 \\
     \textbf{ge2en2ge} &         72.92 &        61.86 &         53.80 &       46.75 \\
\bottomrule
\end{tabular}
    \caption{1,2,3,4-grams {\small BLEU} scores for the paraphrases obtained via back-translation, per language and per language direction.}
    \label{tab:bleu}
\vspace{-0.3cm}
\end{table}

\subsection{Augmented Data: Human Evaluation} \label{app:human}
%back-translations:
%paraphrase type of evaluation
%randomly select 10 instances in one language and the backtranslation from all possible pivot language. e.g. if translation model was available for en into the 5 other languages, then we have 5*10 back to en, so 50 paraphrases to judge.
% we give the reference and up to the 5 paraphrases produced by back-translations and ask annotator to:
% judge fluency, fidelity, surface liability and label preservation
%translations:
% we do not have reference translations, so we can't judge the quality of translation, but rather the quality of the generated paragraphs.
%judge fluency, human-produced or not, label preservation
\paragraph{Data Selection} For \textbf{back-translations}, we randomly extract 10 original input instances in each language, and their back-translations \textit{from} all possible pivot languages, e.g., 10 original \textit{en} instances where each instance was back-translated from \textit{it, fr, ge, es} and \textit{ru}: 10*5 = 50 \textit{en} paraphrases to judge. 
For \textbf{translations}, we randomly extract 10 original input instances in each language and their translations \textit{in} all possible target languages, e.g., 10 original \textit{ge} instances which were translated in \textit{en, fr} and \textit{it}: these 10 translations to be judged in each target language (here \textit{en, fr, it}) originate from the \textit{same} source language. Additionally, in this \textit{translations set}, we add six \textit{control} original instances in each language.  

\paragraph{Annotators} We initially ask five volunteers - one for each language (\textit{en, fr, it, ru, ge}) to partake in the study. One additional annotator (one of this paper's authors) finished annotations for \textit{ge}. All six of them are based in Germany, are native or near-native speaker of the respective language. Each annotator submits two unique sets of answers for (i) translations and (ii) back-translations.

\paragraph{Setup} The annotations are carried out in a remote setting using Google Forms. Annotators are provided detailed written guidelines including examples, first to complete back-translation judgements and then translation judgements ({\small PART 1 \& 2, resp. in Tab.~ \ref{apptab:guidelines}.})
In case of questions, annotators have the option to contact the authors of the paper. 
The evaluation can be completed flexibly in the course of two days. Annotators can take as much as time as they need for completing the evaluation.  
The collected data does not include any information that names or uniquely identifies individual people or offensive content.
Letters of Consent %(cf. §\ref{sec:ethics})
are signed before participation and stored separately from the collected ratings.

\begin{figure*}[!t]
\minipage{0.49\linewidth}
  \includegraphics[width=\linewidth]{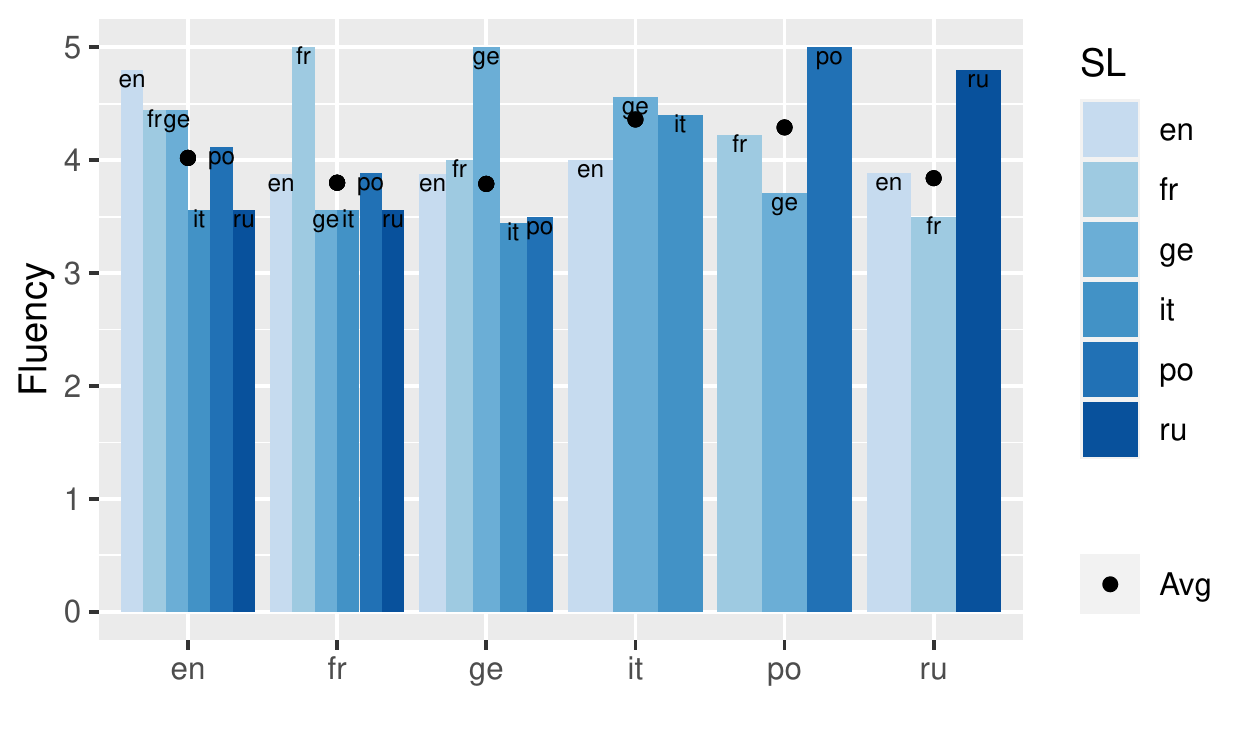}
  \vspace{-0.9cm}
  \caption{Average fluency scores of each {\small TL} (\textit{translations}) irrespective of ({\small \textbf{Avg}}) and according to the source language ({\small \textbf{SL})}.}\label{appfig:T_fluency}
\endminipage\hfill
\minipage{0.49\linewidth}
  \includegraphics[width=\linewidth]{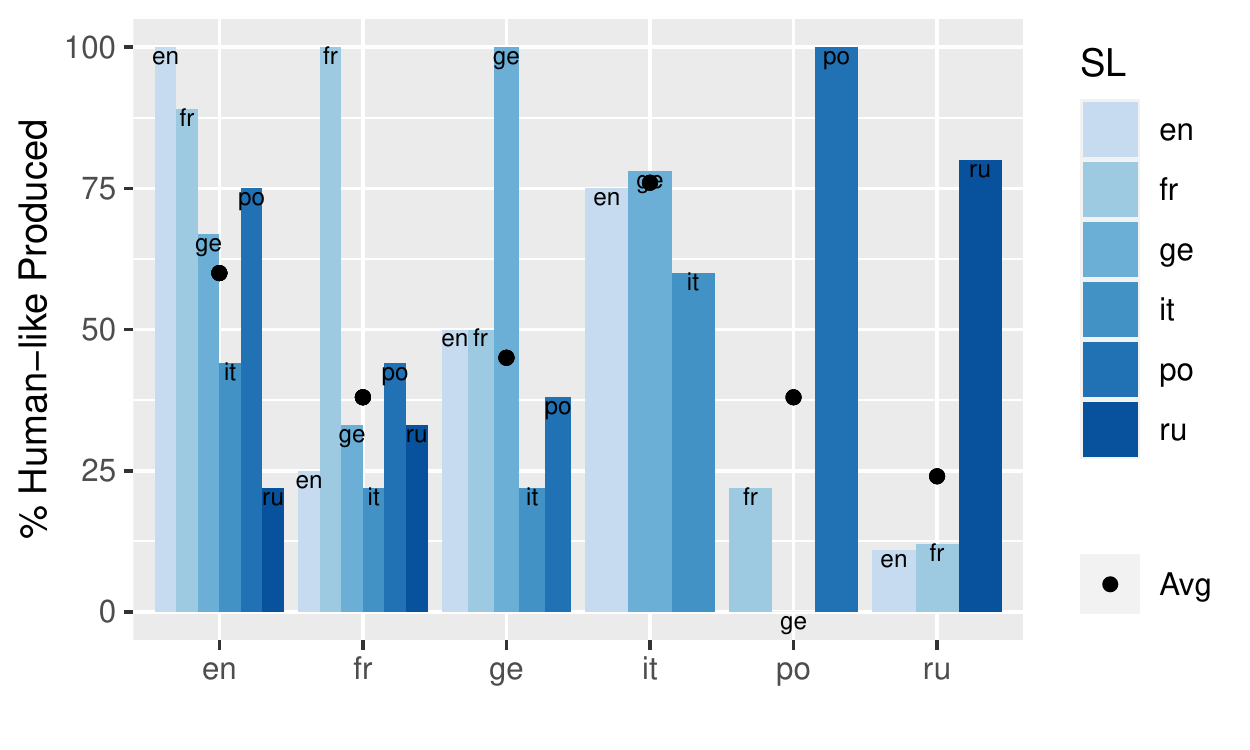}
  \vspace{-0.9cm}
  \caption{Percentage of {\small TL} \textit{translations} judged "human-produced" irrespective of ({\small \textbf{Avg}}) and according to the {\small \textbf{SL}}, respectively.}\label{appfig:T_human_produced}
\endminipage
\end{figure*}
\begin{figure*}[!t]
\minipage{0.32\linewidth}
  \includegraphics[width=\linewidth]{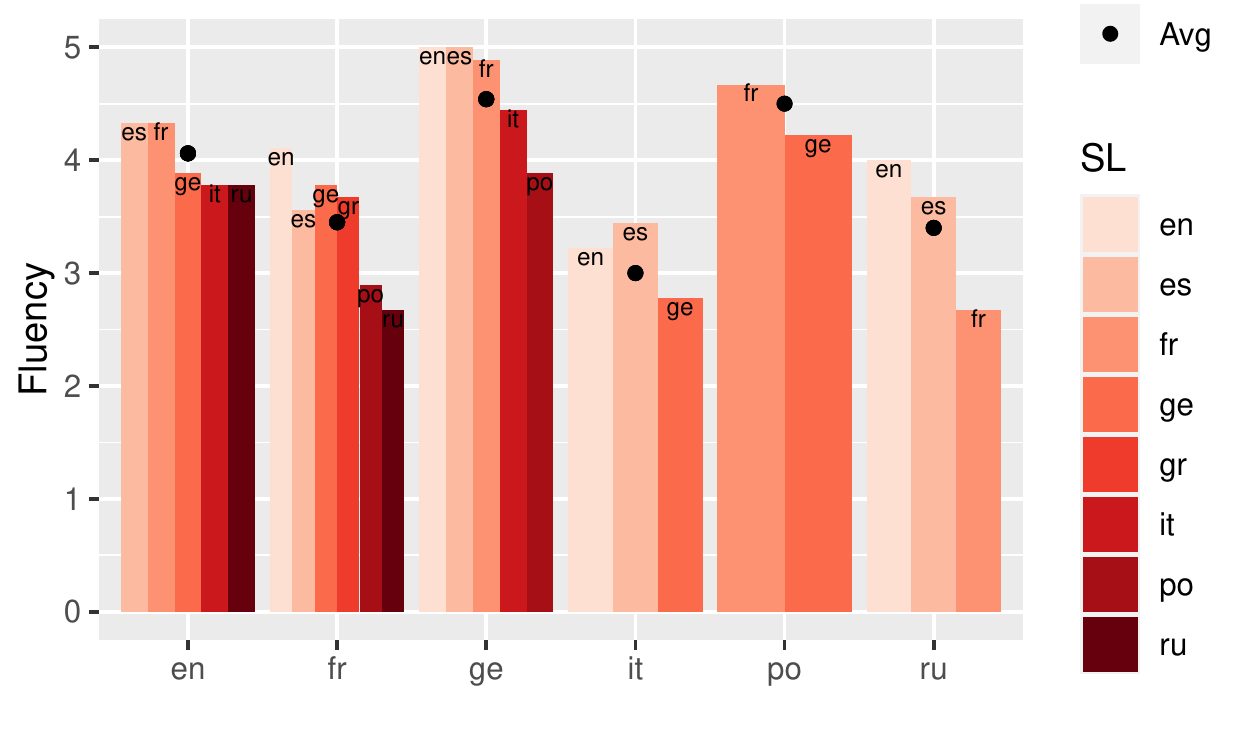}
  \vspace{-0.9cm}
  \caption{Average fluency scores of each {\small TL} (\textit{back-translations}) irrespective of (\textbf{Avg}) and according to the (pivot) (\textbf{SL}).}\label{appfig:BT_fluency}
\endminipage\hfill
\minipage{0.32\linewidth}
  \includegraphics[width=\linewidth]{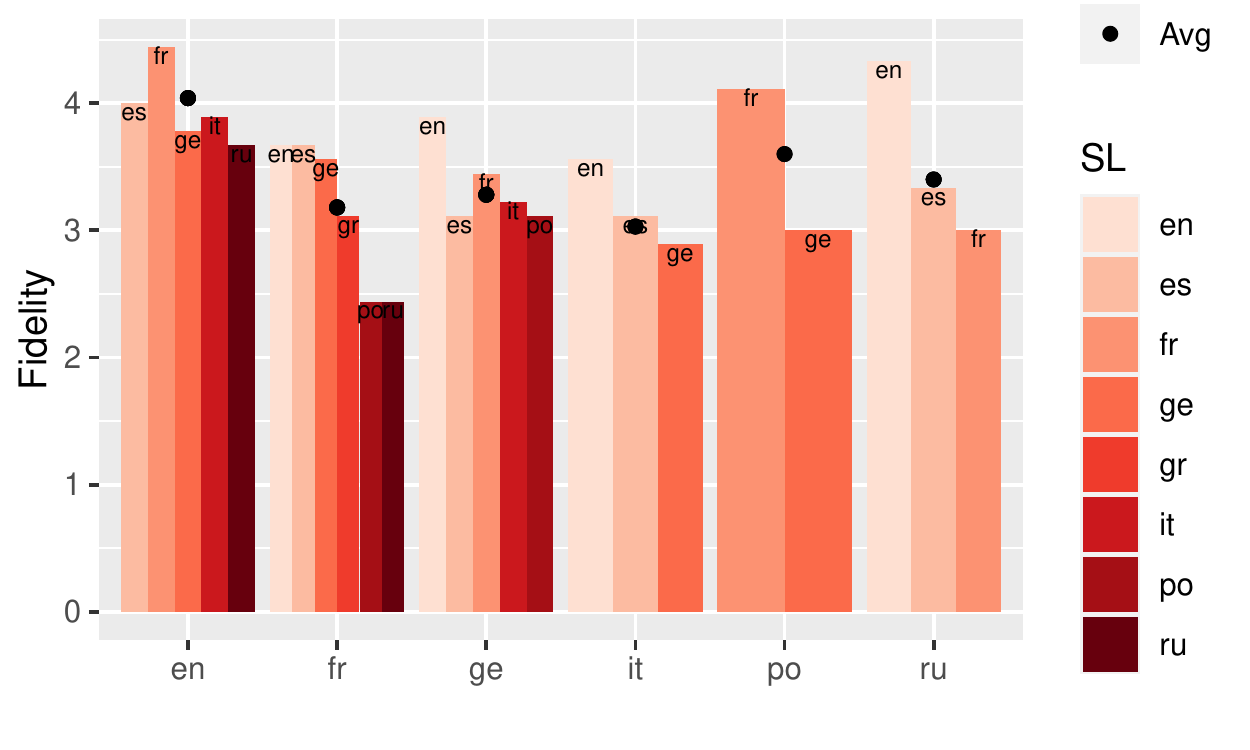}
  \vspace{-0.9cm}
  \caption{Average fidelity scores of each {\small TL} (\textit{back-translations}) irrespective of (\textbf{Avg}) and according to the (pivot) (\textbf{SL}).}\label{appfig:BT_fidelity}
\endminipage\hfill
\minipage{0.32\linewidth}
  \includegraphics[width=\linewidth]{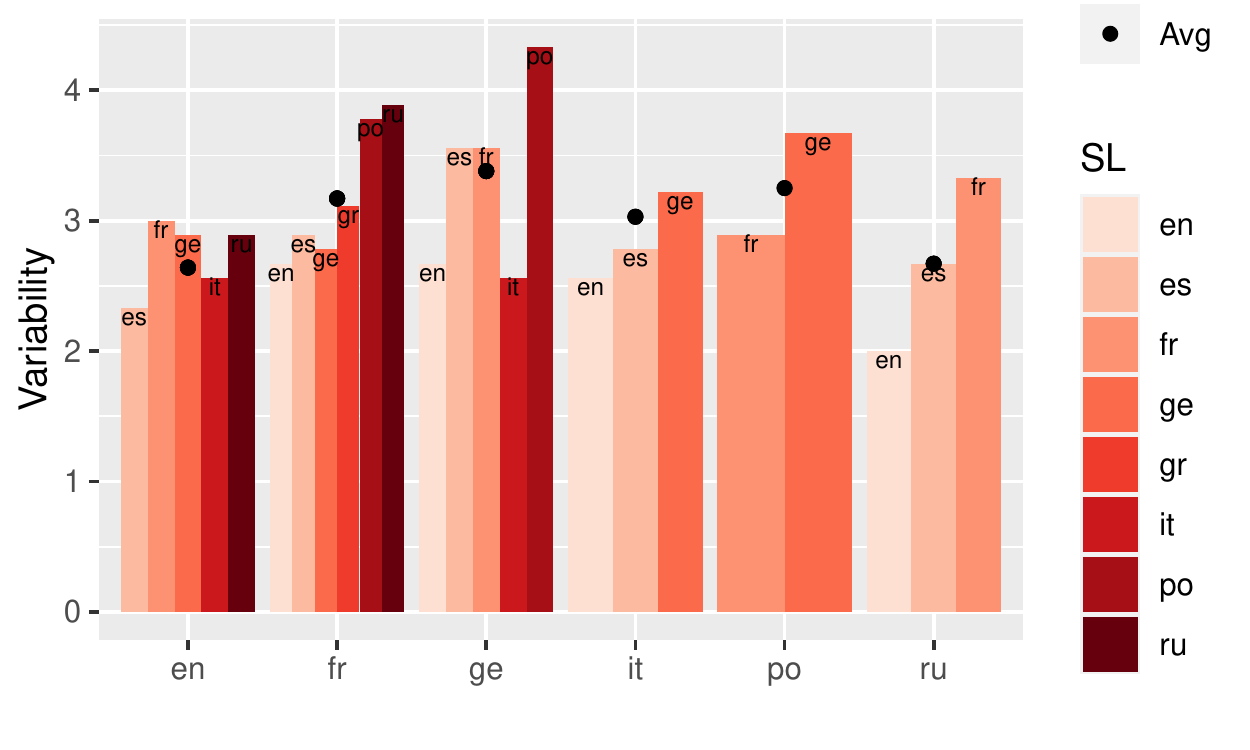}
  \vspace{-0.9cm}
  \caption{Average surface variability scores of each {\small TL} (\textit{back-translations}) irrespective of (\textbf{Avg}) and according to the (pivot) (\textbf{SL}).}\label{appfig:BL_variability}
\endminipage
\end{figure*}

\paragraph{Analyses}
Fig. \ref{appfig:T_fluency} and \ref{appfig:BT_fluency} present the fluency scores attributed to translations and back-translations, respectively. For each target language, we report the average scores irrespective of the source/pivot language (\textbf{Avg}) and the average scores depending on which language it was (back-)translated from. Overall, fluency scores are rather high (avg. 4), which means that (back-)translation does not affect to a large extent the readability of the generated output. However, the percentage of instances that are judged "human-produced" (Fig. \ref{appfig:T_human_produced}) drops with regards to \textit{translations} in \textit{fr, ge} and \textit{ru} (around 30\%). When zooming into language pairs, this percentages drops under 25\% for \textit{en2fr}, \textit{it2fr}, \textit{it2ge} and \textit{{en,fr}2ru}. 
We also collected ratings regarding the \textit{fidelity} and the \textit{surface variability} aspects for the paraphrases obtained via back-translations. On average, and across target languages and language pairs, fidelity scores are found between 3 and 4, which confirms that information gets somewhat lost in back-translation. Regarding surface variability, scores do not go over 3.5: paraphrases do not diverge too much from their respective original inputs but are also not complete copy pastes, which is the desired outcome when paraphrasing. 
We reported in the main text findings regarding the \textit{transfer} and the \textit{preservation} of persuasion techniques in \textit{translation} and \textit{back-translation}, respectively (Fig. \ref{fig:T_loaded_preserved}, \ref{fig:T_slogans_preserved}, \ref{fig:BT_loaded_preserved} and \ref{fig:BT_slogans_preserved} in section \ref{subsec:augmented_data_eval}). We stated that it depends in both cases on the \textit{language pair direction} and the \textit{label} in consideration, but that observations were not always similar for the same \textit{language pair} and \textit{label} depending on translation or back-translation, making the point that certain persuasion techniques may be language- and culture-dependent. From a machine-translation point of view, this finding need to be accounted for when dealing with persuasive text data. 
However, for the translation setup, we had also included six \textit{control} original instances, i.e., in the original language, therefore not translations, to assure that potential label "loss" between all other {\small SL} to {\small TL} was not an artifact of the human judgements. It appears that identifying persuasion techniques, even when provided with a definition and examples, is a difficult task, even for humans: even though these techniques were present in the \textit{gold data}, our annotators judged that \textit{Loaded Language} was "lost" in over 70\% in \textit{ge} of the cases and that \textit{Slogans} was 100\% "lost" in \textit{en} and half the time in \textit{fr} and \textit{ge}.
This raises the limitations of our small-scale annotation study to evaluate the quality of our automatically-obtained (back-)translations. We pave the way for an interesting project regarding the transfer and the preservation of persuasion techniques through (back-)translation, and a larger-scale human evaluation study could be conducted to confirm our findings.

\section{Appendix B}
\subsection{Model Architectures and Hyper-Parameters}\label{subsec:models}
\paragraph{Hyper-Parameter}
Tab.~\ref{tab:hyperparameter} show the hyper-parameter setup for each model architecture. \textit{\#Shots} is the number of instances used to train the SBERT model in a contrastive manner for \texttt{SetFit}. For each instance, 5 triples (original instance, positive, negative) were created.

\paragraph{Libraries}
We use {\small XLM-RoBERTa-base/large} implementations by \texttt{huggingface} \cite{wolf-etal-2020-transformers}, the setfit implementation by \citet{setfit}, and adapaters provided by \citet{pfeiffer2020AdapterHub}. \texttt{scikit-learn} \cite{scikit-learn} is leveraged for SVM and Random Forest implementations as well as pre-processing and metrics. We use the {\small MarianNMT} implementation by \citet{mariannmt} for data augmentation.

%\section{Appendix C}
\subsection{Negative Insights: What did not work} \label{subsec:neg_insights}
Considering {\small XLM-RoBERTa-base} trained on \textit{gold} data as a threshold to be passed, we discard the following models and architectures.

\textbf{Models} SetFit and adapters trained on \textit{gold} only did seriously under-perform our threshold. This observation was one trigger for using data augmentation techniques. Not surprisingly, classic approaches to multi-label classification problem such as {\small SVM}s or Random Forest Classifiers \cite{scikit-learn} did not outperform transformer-based approaches even in the case where more training data was added.  

\textbf{Training Data} Strategically playing around with which relevant training data would lead to better performance was a focus on this work. %was also the To zoom in on relevant training data, 
We showed in Sec. \ref{sec:results} that training with both \textit{gold} and the \textit{spans} increased performance on certain labels. We also tried to train a {\small XLM-RoBERTa-base} model on \textit{spans} only, i.e, discarding the rest of the paragraph's context. %Discarding the context, however, 
This however seriously harmed performance,  %as compared to training on \textit{gold}, 
indicating the importance of larger textual context to detect persuasion techniques. %hence, we decided to train on a concatenation of these two datasets.

To account for label imbalance in the \textit{gold} training data (Tab.~ \ref{tab:label_distribution}), %As the provided \textit{gold} dataset is heavily imbalanced, 
we experiment with injecting translations and/or back-translations only for the paragraphs whose labels are under-represented (< 100) for the respective language, e.g., \textit{Appeal to Time} in \textit{en}, \textit{Repetition} in \textit{ge}.  %for which labels for labels where $n$ training instances $n<$ 100 only. 
Performance drops for most languages, but this is in retrospective not a surprising finding. Not only are persuasion techniques imbalanced \textit{intra-language}, but also \textit{inter-language}: injecting (back-)translations, and therefore transferring persuasion techniques from {\small SL} to {\small TL}, wrongly fills the gaps of label imbalance in the {\small TL}, mistakenly introducing labels that do not \textit{fit} in that {\small TL}.   
%As performance drops for most languages, especially when adding back-translations, we do not consider these models in the ensemble.
% Interestingly, the results were pretty good for ENGLISH. Maybe we could have included them for english only --> and could maybe mention that somewhere (or try it out for camery-ready?) Prisca: I don't think they were good... 10 points less compared to other techniques we tried.
Finally, we also attempt training on languages grouped by their language families (\textit{en-ge}, \textit{fr-it}, \textit{ru-po}), with (i) only the \textit{gold} data and (ii) injecting \textit{translations}. The results vary between languages, but we note improvements on performance, indicating a potential promising direction to take in further experiments.

\section{Appendix C: Quantitative Analysis} \label{sec:regression}

\paragraph{Modeling Setup} We train a {\small XLM-RoBERTa-large} model for 5 epochs on six different training sets, including the \textit{gold} dataset. We measure the {\small F1} score for each of the 23 persuasion strategies on the development set for each language. This results in a data frame of size ($n = 6 \times 23 \times 6 = 828$).

We add a categorical variable as independent variable ({\small IV}) step-by-step, starting with the label. We compare whether the more complex model improves the fit significantly. We then add two-way interactions. 

\begin{figure*}[]
\minipage{0.49\linewidth}
  \includegraphics[width=7.5cm,height=6cm]{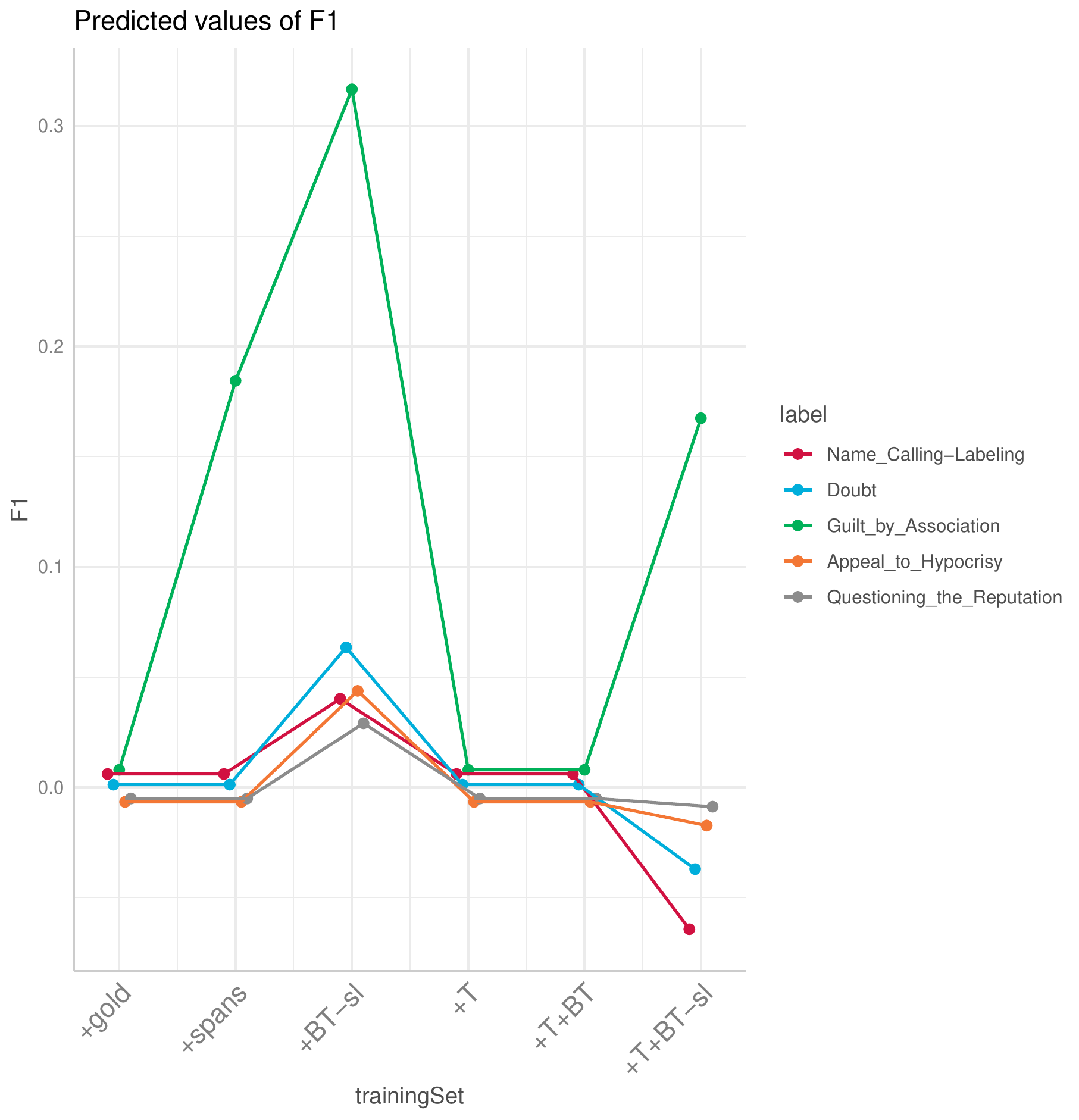}
  %\vspace{-0.9cm}
  \captionsetup{font=footnotesize}
  \caption{Effect of training data type (+size, in ascending order left to right) on predicted {\small F1} scores for persuasion techniques falling under \textit{Attack of Reputation}.}\label{appfig:attack}
\endminipage\hfill
\minipage{0.49\linewidth}
  \includegraphics[width=7.5cm,height=6cm]{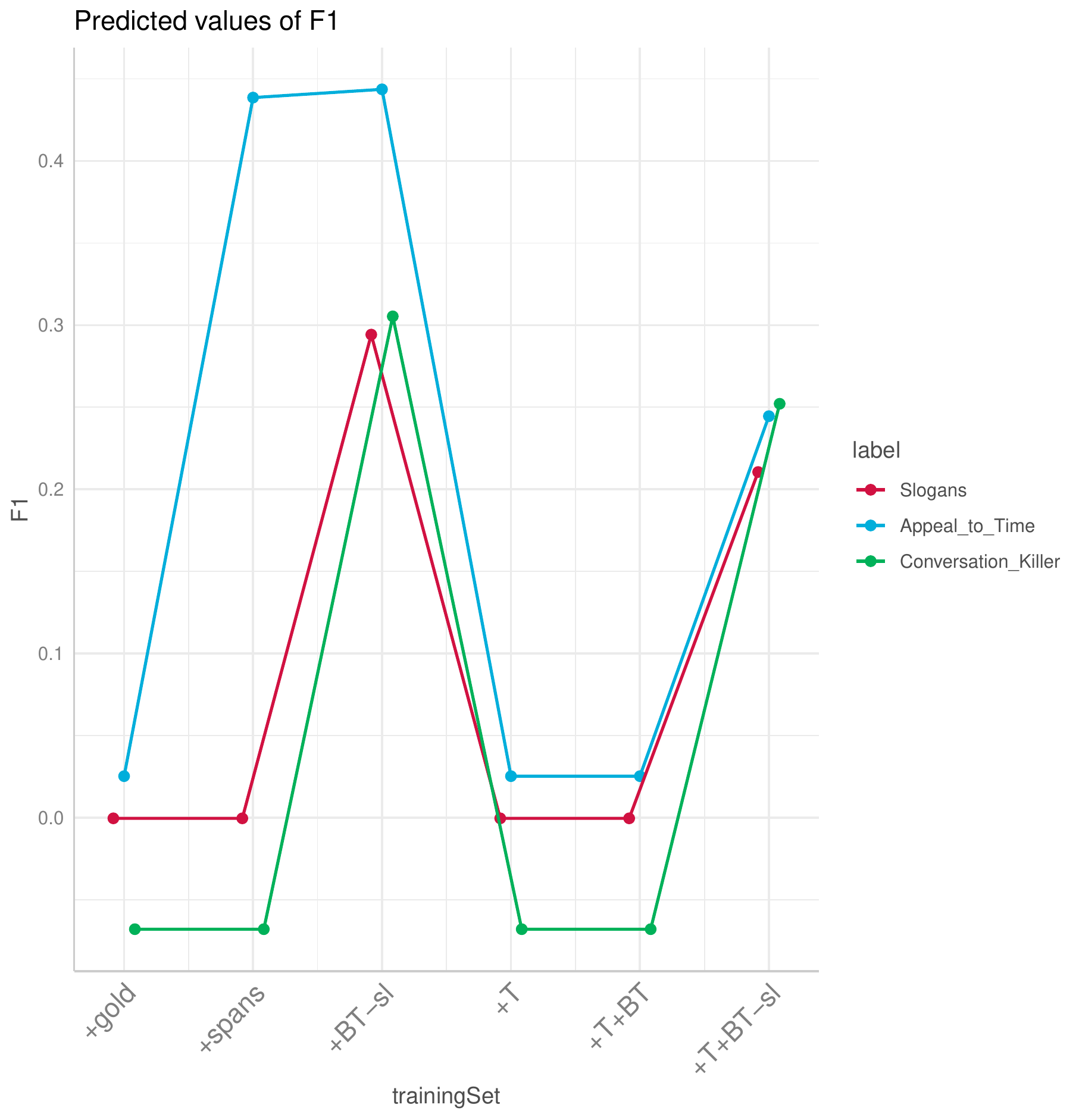}
  %\vspace{-0.9cm}
  \captionsetup{font=footnotesize}
  \caption{Effect of training data type (+size, in ascending order left to right) on predicted {\small F1} scores for persuasion techniques falling under \textit{Call}.}\label{appfig:call}
\endminipage
\end{figure*}

\begin{figure*}[]
\minipage{0.49\linewidth}
  \includegraphics[width=7.5cm,height=6cm]{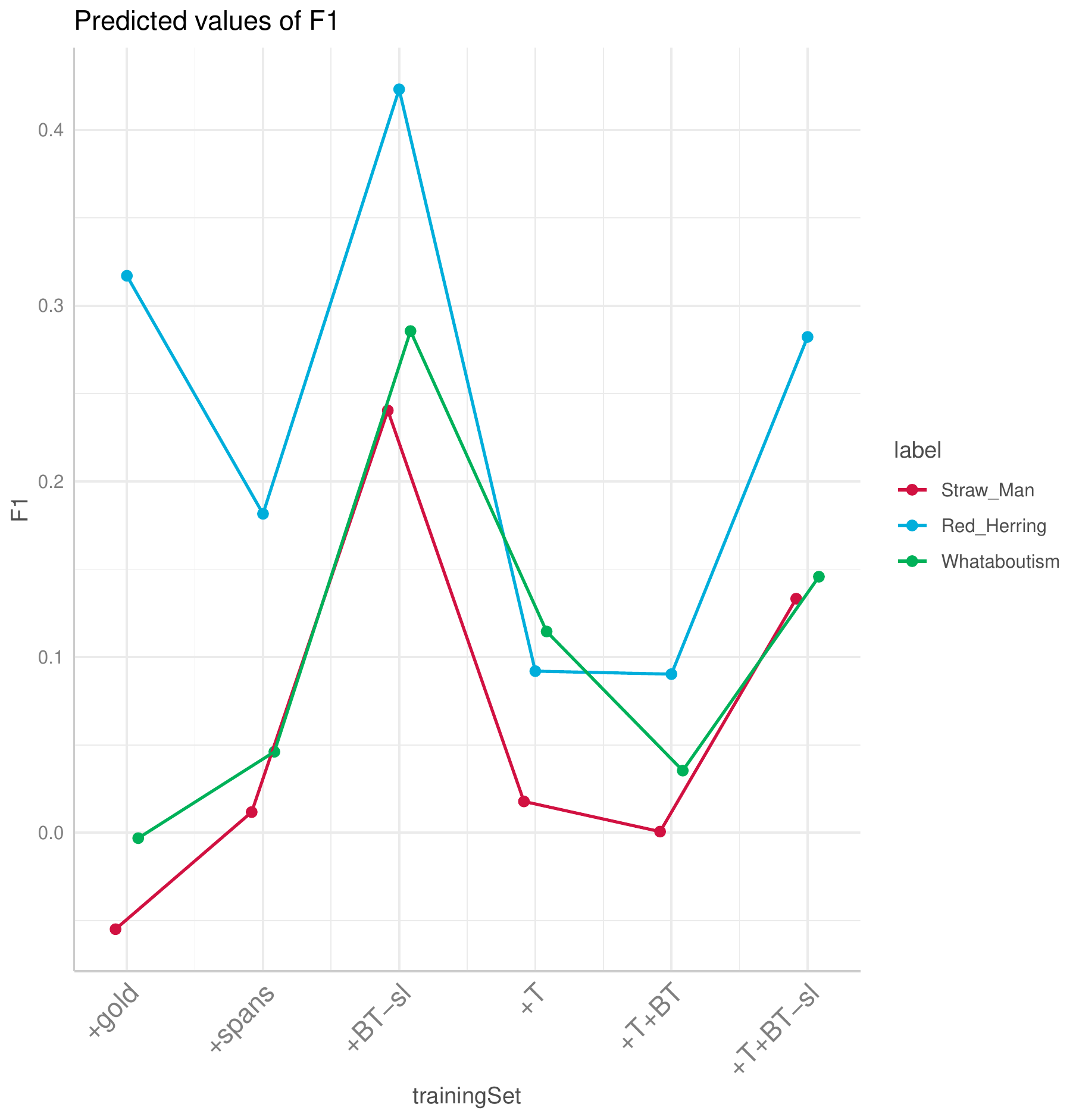}
  %\vspace{-0.9cm}
  \captionsetup{font=footnotesize}
  \caption{Effect of training data type (+size, in ascending order left to right) on predicted {\small F1} scores for persuasion techniques falling under \textit{Distraction}.}\label{appfig:distraction}
\endminipage\hfill
\minipage{0.49\linewidth}
  \includegraphics[width=7.5cm,height=6cm]{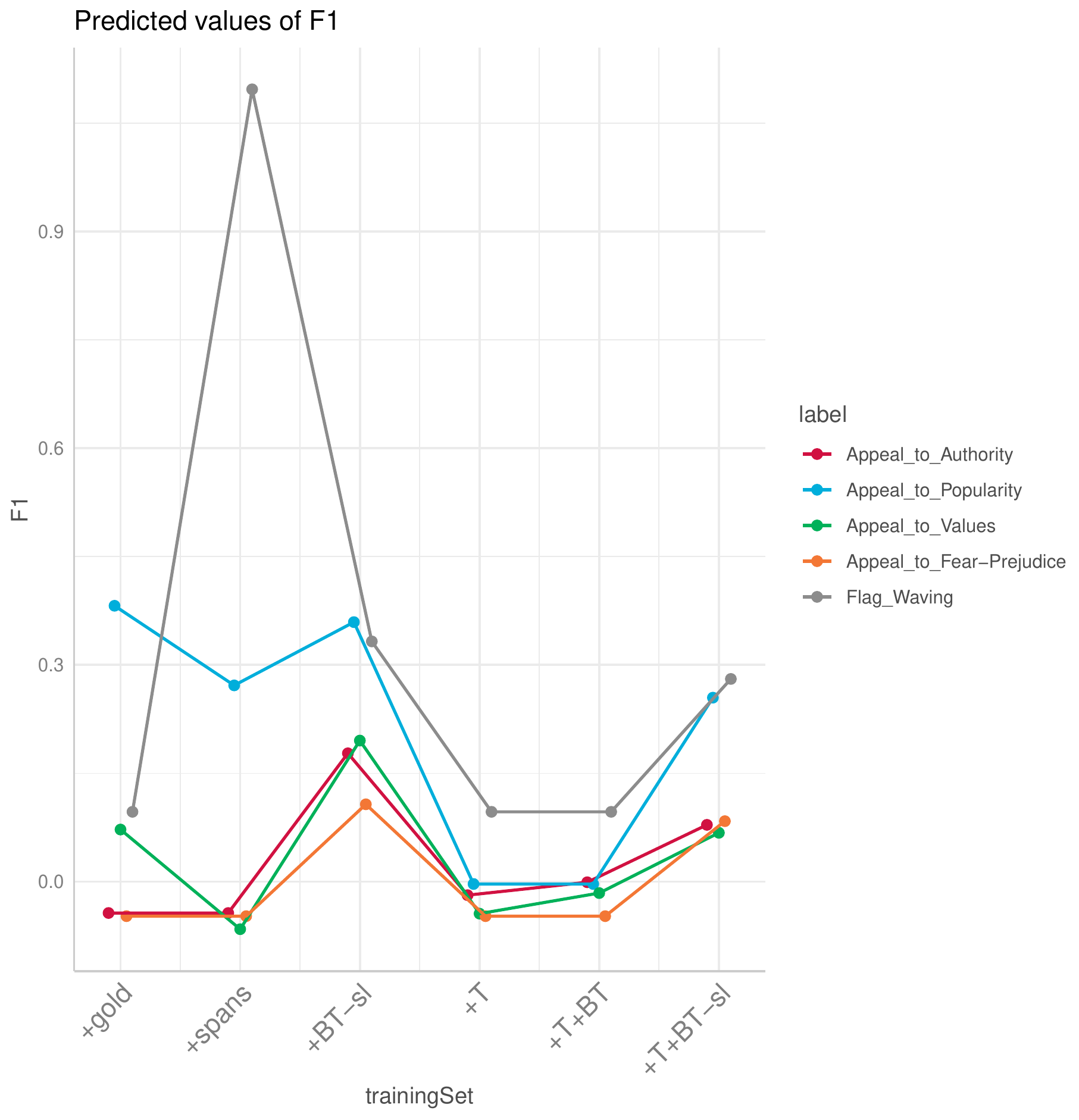}
  %\vspace{-0.9cm}
  \captionsetup{font=footnotesize}
  \caption{Effect of training data type (+size, in ascending order left to right) on predicted {\small F1} scores for persuasion techniques falling under \textit{Justification}.}\label{appfig:justification}
\endminipage
\end{figure*}

\begin{figure*}[]
\minipage{0.49\linewidth}
  \includegraphics[width=7.5cm,height=6cm]{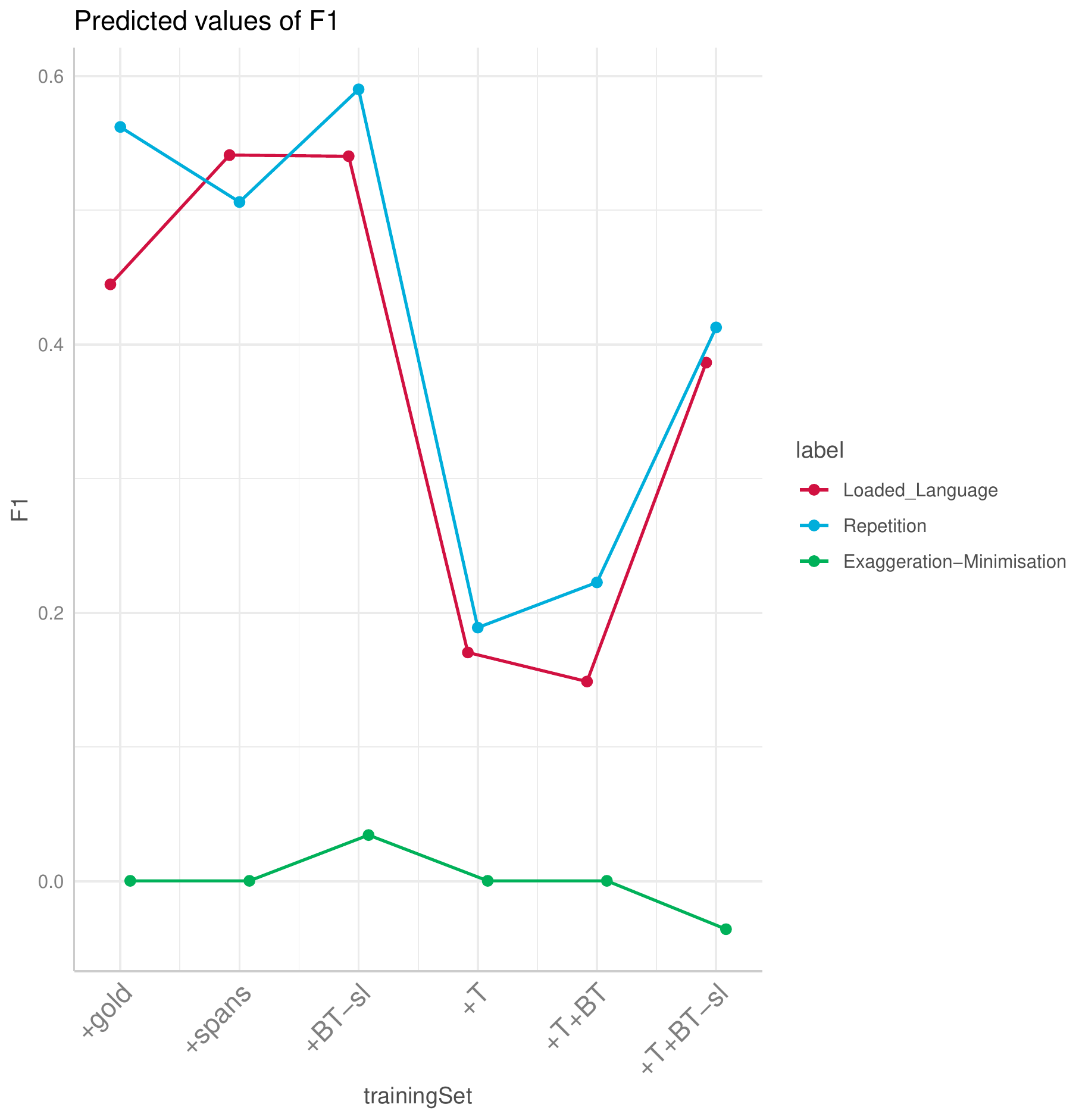}
  %\vspace{-0.9cm}
  \captionsetup{font=footnotesize}
  \caption{Effect of training data type (+size, in ascending order left to right) on predicted {\small F1} scores for persuasion techniques falling under \textit{Manipulative Wording}.}\label{appfig:manipulative}
\endminipage\hfill
\minipage{0.49\linewidth}
 \includegraphics[width=7.5cm,height=6cm]{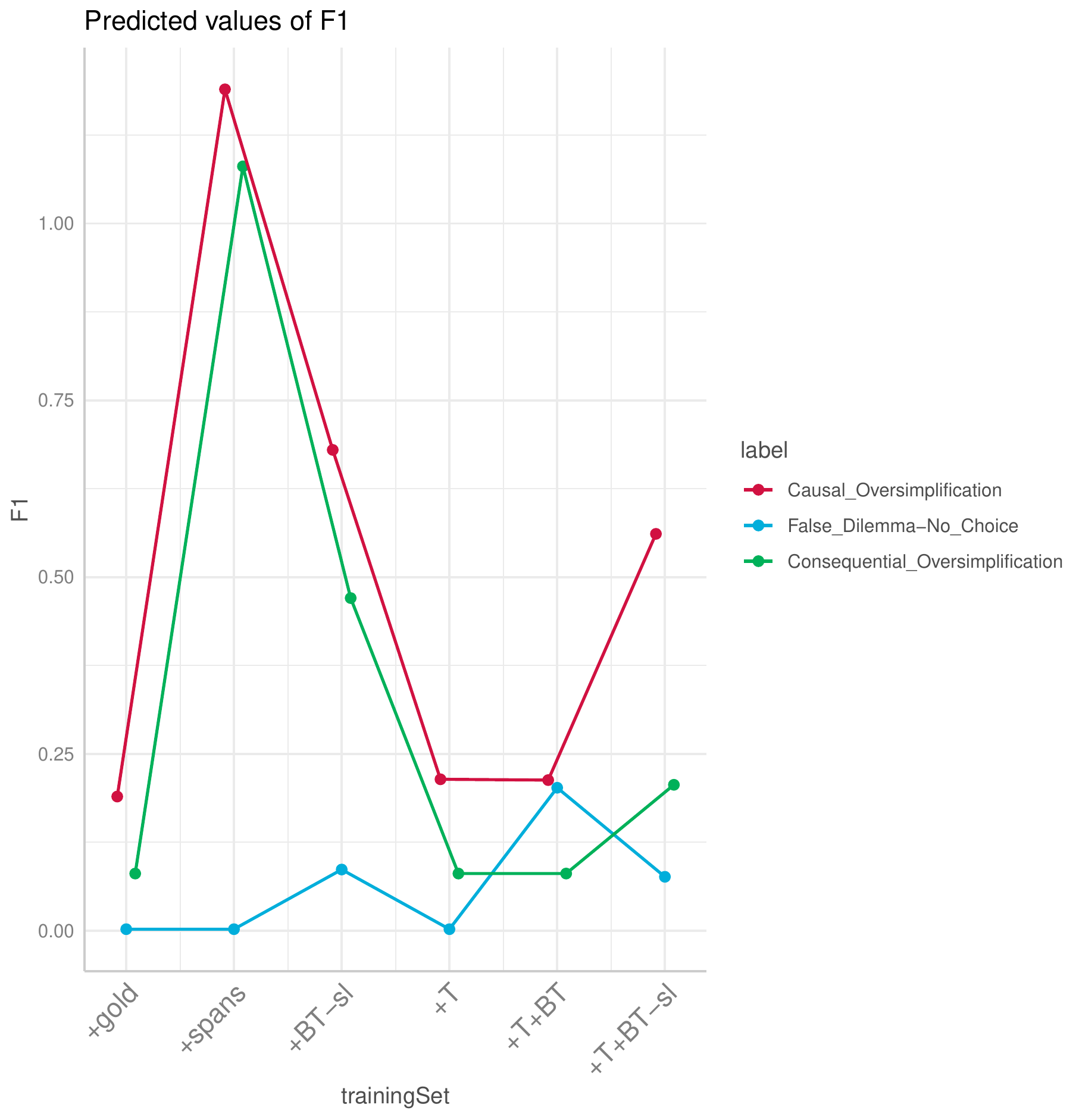}
  %\vspace{-0.9cm}
 \captionsetup{font=footnotesize}
 \caption{Effect of training data type (+size, in ascending order left to right) on predicted {\small F1} scores for persuasion techniques falling under \textit{Simplification}.}\label{appfig:simplification}
\endminipage
\end{figure*}

\paragraph{Analyses} We presented in Sec.~\ref{sec:results} the effect of training data on prediction scores for six persuasion techniques (Fig.~\ref{interaction_corpus_label}). We report this effect on \textit{all} 
persuasion techniques, from Fig.~\ref{appfig:attack}--\ref{appfig:simplification}. 
Similarly to our findings on six techniques, results clearly point to back-translation ({\small +BT-sl}) as the most robust augmentation strategy with consistent improvement of {\small F1} scores across all labels.

\begin{table*}[b]
    \centering
    \footnotesize
    \begin{tabular}{l}
    \toprule
    \textbf{PART 1}: You will evaluate the \textbf{quality of generated paraphrases.} \\\midrule
    You will be given the original sentence and several paraphrases for that original sentence. For each paraphrase: \\
    \\
    1) On a scale from 1 to 5, rate the \textbf{fluency} of that paraphrase. \\
    Irresp. of the original sentence, how readable is the paraphrase? \\
    1 means the sentence is not readable/plausible at all, and 5 is fully fluent. \\
    \\
    2) On a scale from 1 to 5, rate the \textbf{fidelity} of that paraphrase. \\
    Compared to the original sentence, how much information is preserved? -\\
    How semantically consistent is the paraphrase? \\
    1 is when the information is fully lost, and 5 is fully semantically consistent. \\
    \\
    3) On a scale from 1 to 5, rate the \textbf{surface variability} of that paraphrase.\\ How much difference does the paraphrase have in the form of expression compared to the original sentences?\\
    1 is a word-per-word paraphrase, and 5 is a fully new constructed sentence. \\
    \\
    4) A \textbf{persuasion technique} is assigned to the original sentence: \\
    Does it apply to the paraphrase as well? Yes or No.\\
    \\
    We show an example below. \\
    You will encounter only two different persuasion techniques: \textbf{Loaded\_Language} and \textbf{Slogans}.\\
    We provide below a short definition and a couple of examples in English for you to get an overall idea of the techniques. \\
    \\
    \includegraphics[width=16cm]{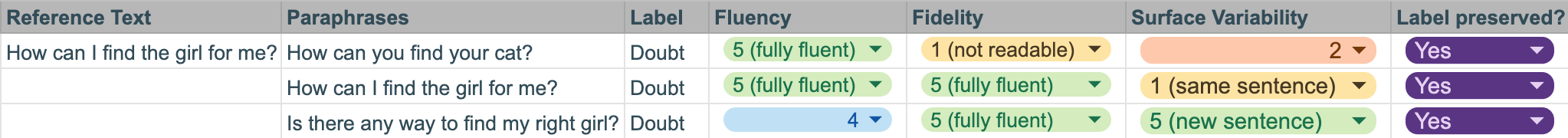}
    \\\midrule \\\midrule
    \textbf{Loaded\_Language:} Using specific words and phrases with strong emotional implications (either positive or negative) to \\
    influence and convince the audience that an argument is valid/true. \\
    This fallacy is also known as euphemisms, appeal to/argument from emotive language, or loaded language.\\
    \\
    \underline{Examples: }\\
    - “How \textbf{\textit{stupid}} and \textbf{\textit{petty}} things have become in Washington” \\
    - “They keep feeding these people with \textbf{\textit{trash}}. They should stop.” \\
    \\
    \textbf{Slogans}: A brief and striking phrase that may include labeling and stereotyping. Slogans tend to act as emotional appeals.\\
    \\
    \underline{Examples:}\\ 
    - “Our \textbf{\textit{'unity in diversity'}} contrasts with the divisions everywhere else.” \\
    - \textbf{\textit{“Make America great again!”}} \\
    - \textbf{\textit{“Immigrants welcome, racist not!”, “No border. No control!”}} \\\midrule
     \\\midrule
    \textbf{PART 2}: You will evaluate the \textbf{quality of generated paragraphs}.\\\midrule
    You will be given sentences, for each of them:\\
    \\
    1) On a scale from 1 to 5, rate the \textbf{fluency} of that paraphrase: \\
    How readable is the paraphrase?\\
    1 means the sentence is not readable/plausible at all, and 5 is fully fluent.\\
    \\
    2) Do you think this sentence was \textbf{human-produced} (vs. automatically generated)? Yes or No. \\
    3) A \textbf{persuasion technique} is assigned to the sentence:\\
    Do you think it fits? Yes or No.
    \\
    We show an example below.
    \\
    If needed, we provide the info on the persuasion techniques we consider once again (above in this paper). \\
    \\
    \includegraphics[width=13cm]{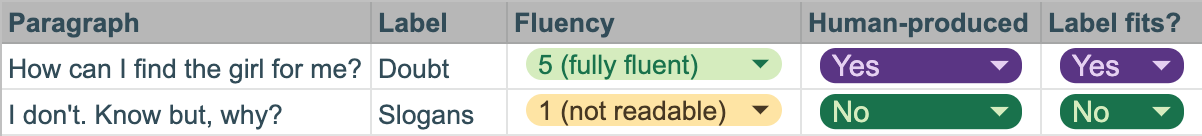} \\\bottomrule        
    \end{tabular}
    \caption{Annotation guidelines for the human annotation study.}
    \label{apptab:guidelines}
\end{table*}

\begin{table*}[!ht]
\small
\centering
\begin{tabular}{@{}lllll@{}}

\toprule
                      & XLM-R-base           & XLM-R-large          & Adapters             & SetFit            \\ \midrule
learning rate         & 1e-5                 & 1e-5                 & 1e-4                 & -                 \\
max epochs            & 10                   & 5                    & 5                    & 2                 \\
num of text pairs     & -                    & -                    & -                    & 5                 \\
\#shots               & -                    & -                    & -                    & 1000              \\
seed                  & 42                   & 42                   & 42                   & 42                \\
batch size            & 16                   & 16                   & 16                   & 32                \\
loss                  & Binary Cross Entropy & Binary Cross Entropy & Binary Cross Entropy & Cosine Similarity \\
metric for best model & F1 macro             & F1 macro             & F1 macro             & F1 macro          \\ \bottomrule
\end{tabular}
\caption{Overview of hyper-parameter for each model architecture used for the submission on the final test set.}
\label{tab:hyperparameter}
\end{table*}

\end{document}